\documentclass{article}

\usepackage[preprint]{neurips_2026}

\usepackage[utf8]{inputenc}
\usepackage[T1]{fontenc}
\usepackage[breaklinks=true]{hyperref}
\usepackage{url}
\usepackage{booktabs}
\usepackage{multirow}
\usepackage{placeins}
\usepackage{amsfonts}
\usepackage{amsmath}
\usepackage{amssymb}
\usepackage{graphicx}
\usepackage{nicefrac}
\usepackage{microtype}
\usepackage{wrapfig}
\usepackage{xcolor}
\usepackage[most]{tcolorbox}
\usepackage{enumitem}
\usepackage{capt-of}
\usepackage{listings}
\usepackage{adjustbox}
\usepackage{hyperref}

\title{CodeTS: Verifiable Text-to-Time Series Generation via Executable Code}
\author{
\mdseries
Xudong Yuan$^{1}$ \,
Shunyu Liu$^{2}$ \,
Tongya Zheng$^{1}$ \,
Huiping Zhuang$^{3}$ \,
Mingli Song$^{1}$ \,
Kaixuan Chen$^{1}$ \\
$^{1}$Zhejiang University, Hangzhou, China \\
$^{2}$Nanyang Technological University, Singapore \\
$^{3}$South China University of Technology, Guangzhou, China
}

\begin{document}

\maketitle

\begin{abstract}
Text-to-Time Series Generation (Text-to-TS) provides a promising paradigm for synthesizing time series from natural language, enabling scenario-specific generation when real observations are scarce or costly to acquire.
However, existing methods typically lack an explicit mechanism for deriving generation logic from textual descriptions to guide time series synthesis.
In this paper, we propose CodeTS, a verifiable framework that uses code as an intermediate generation interface, reformulating Text-to-TS generation as a Text-to-Code-to-TS process.
CodeTS first maps textual temporal descriptions into an explicit code space, where executable code specifies how textual requirements shape target temporal patterns, and then obtains the time series through code execution.
To learn this code generation process reliably without real code annotations, CodeTS constructs aligned Text-Code-TS triplets from structured temporal attributes for supervised initialization.
More importantly, we further design multi-stage execution-based rewards that verify format validity, code executability, and time series quality, enabling real Text-TS pairs to provide training signals for Reinforcement Learning with Verifiable Rewards (RLVR).
Extensive experiments on eight benchmarks across short, medium, and long generation lengths demonstrate that CodeTS provides a strong zero-shot solution for Text-to-TS generation, outperforming LLM-based baselines and achieving better averaged results than supervised generative baselines trained on the target datasets.
% Extensive experiments on eight benchmarks across short, medium, and long generation lengths demonstrate that CodeTS achieves state-of-the-art zero-shot Text-to-TS generation performance and consistently outperforms strong supervised and LLM-based baselines.
The anonymized code repository is available at \url{https://anonymous.4open.science/r/CodeTS-54DA}.
\end{abstract}

\section{Introduction}

\begin{wrapfigure}{r}{0.5\textwidth}
\vspace{-1.2em}
\centering
\setlength{\abovecaptionskip}{2pt}
\setlength{\belowcaptionskip}{0pt}
\includegraphics[width=\linewidth,trim=8pt 8pt 8pt 8pt,clip]{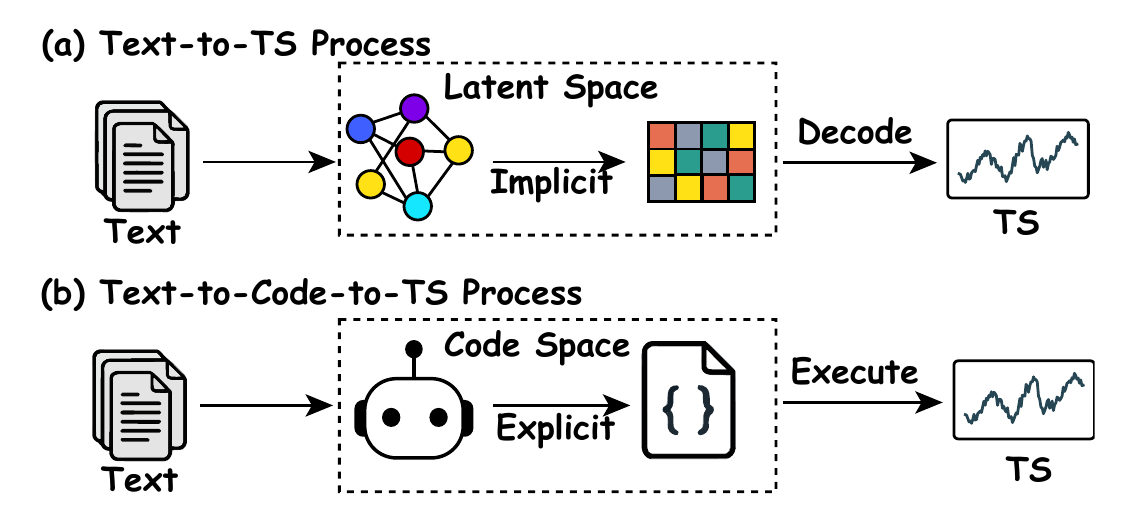}
\caption{Illustration of the Text-to-TS process and the proposed Text-to-Code-to-TS process.}
\label{fig:illustration}
\vspace{-0.8em}
\end{wrapfigure}

Time series data are pervasive~{across real world domains} such as finance, healthcare,~{energy systems, transportation networks, and climate science}~\citep{SEZER2020106181,faust2018deep,DEB2017902,ERMAGUN2018786,AN2025126301}, where they support fundamental tasks including forecasting, classification, and anomaly detection~\citep{wen2022transformers}.~{These tasks require representative data that adequately cover diverse temporal patterns and characteristic behaviors across application contexts. However, such comprehensive time series data are often scarce and costly to obtain, especially when the target patterns involve rare events, privacy restricted sensor measurements, or expert descriptions of temporal behaviors} \citep{ge2025t2s,wu2025scits,wen2021time}.~{As a result, many important temporal patterns remain insufficiently covered, limiting model performance in downstream scenarios where such patterns are most critical. Therefore, this underscores the need for time series generation methods that synthesize samples with targeted temporal characteristics rather than simply increasing the number of synthetic samples.}

Existing time series generation methods have mainly focused on learning temporal distributions from observed series and synthesizing new samples with similar statistical properties~\citep{jeong2025frequency,yoon2019time,lee2023vector,yuan2024diffusionts}. 
To improve controllability, conditional generation methods further incorporate auxiliary signals, such as metadata, labels, or historical observations, to guide the synthesis process~\citep{narasimhan2024time,shankar2025wavestitch,lin2019doppelganger,ni2020conditional,liao2024sig}. 
% However, these forms of conditioning are often predefined and structured, making it difficult to express fine-grained temporal requirements, such as desired trends, periodicity, amplitude changes, and local events, in a flexible and intuitive manner. 
However, these forms of conditioning are often predefined and structured, making it difficult to specify fine-grained temporal requirements \citep{gu2025verbalts}, such as desired trends, periodicity, amplitude changes and local events, in a flexible and intuitive manner. 
Natural language, in contrast, provides a more expressive interface for specifying such target temporal behaviors, thereby motivating time series generation from text.

Following this direction, Text-to-Time Series Generation (Text-to-TS) has emerged as a promising paradigm for natural language-guided time series synthesis, enabling compositional temporal requirements to be specified in a flexible form. 
Existing approaches can be broadly divided into supervised Text-to-TS generation methods and LLM-based generation methods.
The former learn Text-to-TS mappings from paired Text-TS data, typically by aligning textual and temporal representations and training neural generators to synthesize numerical sequences~\citep{gu2025verbalts,ge2025t2s,li2025bridge,zhang2026spectral}; 
the latter use large language models either through prompting or by autoregressively modeling textualized time series representations to synthesize temporal data~\citep{rousseau2025forging,xie2025chatts,wang2025chattime,guan2025timeomni}.
However, these existing methods connect textual descriptions and temporal patterns through latent representation spaces (Fig.~\ref{fig:illustration}(a)), where the transformation from textual semantics to temporal patterns remains implicit and underspecified, which leaves the Text-TS semantic gap insufficiently addressed and makes the underlying generation logic difficult to verify or optimize. This raises a key challenge: 
\begin{tcolorbox}[
  enhanced,
  colback=blue!1,
  colframe=blue!35!black,
  boxrule=0.45pt,
  arc=1.5mm,
  left=6pt,
  right=6pt,
  top=5pt,
  bottom=5pt,
  before skip=6pt,
  after skip=6pt
]
% \emph{How to develop an explicit intermediate interface that translates textual requirements into time series generation logic, while making the translation verifiable and optimizable to better bridge the semantic gap between language and temporal dynamics?}
\emph{How to develop an explicit intermediate interface that maps textual requirements into time series generation logic, while making the process verifiable and optimizable to better bridge the semantic gap between textual descriptions and temporal dynamics?}
\end{tcolorbox}

% In this paper, we propose CodeTS, a novel Text-to-Code-to-TS framework that bridges the semantic gap between textual descriptions and temporal dynamics by introducing executable code as an explicit intermediate interface, as illustrated in Fig.~\ref{fig:illustration}(b). Specifically, CodeTS decomposes Text-to-TS generation into Code Generation (CodeGen), which grounds textual temporal requirements into executable generation programs, and Code Execution (CodeExe), which executes these programs to produce time series. 
% This decomposition transforms the implicit language-to-dynamics mapping into an explicit and executable generation process, enabling verification through execution while allowing the code generation policy to be initialized with supervised learning and further optimized through reinforcement learning.

In this paper, we propose CodeTS, a novel Text-to-Code-to-TS framework that bridges the semantic gap between textual descriptions and temporal dynamics by introducing executable code as an explicit intermediate interface, as illustrated in Fig.~\ref{fig:illustration}(b). Specifically, CodeTS decomposes Text-to-TS generation into Code Generation (CodeGen), which grounds textual temporal requirements into executable generation programs, and Code Execution (CodeExe), which executes these programs to produce time series. 
This decomposition transforms the implicit Text-to-TS mapping into an explicit generation process, enabling verification through execution while allowing the code generation policy to be initialized with supervised learning and further optimized through reinforcement learning.
To support this two-stage learning process, we construct aligned Text-Code-TS triplets from structured temporal attributes for supervised initialization, and design multi-stage execution-based rewards over format validity, code executability, and time series quality for reinforcement learning optimization. 
Finally, extensive experiments on eight benchmarks across different generation lengths demonstrate the effectiveness of CodeTS compared with state-of-the-art baselines.

Our main contributions are summarized as follows:
\begin{itemize}[leftmargin=2.6em, itemsep=0.2em, parsep=0pt, topsep=0.1em, partopsep=0pt]
    \item 
    We are the first to reformulate Text-to-TS generation as a Text-to-Code-to-TS process, transforming the implicit mapping between text and time series into an explicit code space. This formulation provides a new perspective for mitigating the gap between language semantics and temporal dynamics.

    \item 
    We propose CodeTS, a novel framework that implements Text-to-Code-to-TS with normalized code generation and sandboxed execution, making time series generation explicit, verifiable, and optimizable via execution-based rewards.
    
    % We propose CodeTS, a novel framework that implements the Text-to-Code-to-TS formulation with normalized code generation and sandboxed code execution.
    % By grounding temporal factors such as trend, seasonality, events, and noise into programmatic operations, CodeTS makes the generation process explicit and verifiable through execution, and optimizable via execution-based rewards.
    
    % We propose CodeTS, a novel framework that grounds temporal factors into executable programs, making Text-to-TS generation explicit, verifiable, and optimizable via normalized code generation, sandboxed execution, and execution-based rewards.

    \item  
    We develop aligned Text-Code-TS triplets for supervised initialization and multi-stage execution rewards for reinforcement learning.
    These designs provide code-level supervision and multi-stage verification over format validity, code executability, and time-series quality.

    % We design aligned Text-Code-TS triplets for supervised initialization and multi-level execution rewards for reinforcement learning optimization.
    % The triplets provide code-level supervision, while the rewards enable multi-level verification over format validity, code executability, and time-series quality.

    % We develop a two-stage learning scheme for CodeTS, with supervised initialization from constructed Text-Code-TS triplets and execution-based reinforcement learning on real Text-TS pairs.
    % Multi-level execution rewards evaluate format validity, code executability, and time-series quality, enabling policy refinement with GRPO.

    % We develop a two-stage learning scheme for CodeTS, including supervised initialization from constructed Text-Code-TS triplets and execution-guided reinforcement learning optimization on real Text-TS pairs. 
    % Multi-level execution-based rewards evaluate format validity, code executability, and time-series quality, enabling effective policy refinement with RLVR and GRPO.
    
    \item 
    Extensive experiments on eight benchmarks across different generation lengths demonstrate the effectiveness of CodeTS compared with state-of-the-art supervised Text-to-TS methods and LLM-based zero-shot baselines.

\end{itemize}

\section{Related Work}

\textbf{Time Series Generation and Text-to-TS Generation.}
Deep generative models for time series have progressed from GAN-based approaches, such as TimeGAN~\citep{yoon2019time} and GT-GAN~\citep{jeon2022gtgan}, to diffusion and discrete-latent models, such as Diffusion-TS~\citep{yuan2024diffusionts} and TimeVQVAE~\citep{lee2023vector}. Recent Text-to-TS studies extend conditional generation to natural language, including VerbalTS~\citep{gu2025verbalts}, T2S~\citep{ge2025t2s}, BRIDGE~\citep{li2025bridge}, and textualized autoregressive generation methods~\citep{rousseau2025forging}. Despite their differences in supervision, architecture, and generation strategy, existing methods mostly rely on implicit representations, making the gap between textual requirements and temporal patterns difficult to inspect and directly optimize.

\textbf{Language Models and Foundation Models for Time Series.}
Recent studies have explored language models and foundation models for time series forecasting, representation learning, and multimodal understanding. Representative examples include LLM-based forecasting methods such as GPT4TS~\citep{zhou2023one} and Time-LLM~\citep{jin2024time}, foundation models such as Chronos~\citep{ansari2024chronos}, and multimodal time series systems such as ChatTime~\citep{wang2025chattime} and TimeOmni-1~\citep{guan2025timeomni}. 
These works highlight the potential of language model-based time series modeling, but still rely on numerical, discretized, or textualized sequence representations, leaving the translation from textual requirements to temporal patterns largely implicit.
Additionally, TS2Code~\citep{tan2026ts2code} recently explores code-based time series understanding by generating code from time series visualizations, but it focuses on reconstruction and forecasting rather than text-conditioned generation.

\section{Problem Definition}
Text-to-TS generation aims to synthesize a time series conditioned solely on a natural language description. Following the setting in~\citep{ge2025t2s}, the dataset with \(N\) samples can be denoted as:
\begin{equation}
\mathcal{D}
=
\{(x_i, d_i)\}_{i=1}^{N},
\label{eq:text_series_dataset}
\end{equation}
where \(x_i \in \mathbb{R}^{L_i}\) denotes a univariate time series of length \(L_i\) and \(d_i\) denotes its corresponding description.
The generated time series \(\hat{x}_i\) is expected to match the target time series \(x_i\), which can be written as:
\begin{equation}
\min_{\theta}\;
\mathbb{E}_{(x_i, d_i)\sim\mathcal{D}}
\big[ \| x_i - \hat{x}_i \|^2_2 \big],
\qquad
\hat{x}_i \sim \pi_\theta(\cdot \mid d),
\label{eq:text_to_ts_objective}
\end{equation}
where \(\pi_\theta\) denotes a conditional generative model. In this formulation, \(\pi_\theta\) can be instantiated as a text-conditioned diffusion or as an LLM-based generator. However, the generation logic that maps textual semantics to temporal dynamics remains implicit.

% where \(d_i\) is a natural language temporal description associated with the  time series \(x_i\) and encodes its global and local temporal properties, such as trend, periodicity, local events, amplitude changes, and noise characteristics. 
\section{Methodology}
This section presents CodeTS, which reformulates Text-to-TS generation as Text-to-Code-to-TS: a textual description is first translated into executable code, and the time series is then obtained by code execution. 
We first formalize this formulation in Sec.~\ref{sec:formulation}, then describe how aligned Text-Code-TS triplets are constructed to provide code-level supervision in Sec.~\ref{sec:triplet}. 
We next introduce execution-based rewards for evaluating generated code and its executed time series in Sec.~\ref{sec:reward}, followed by the supervised initialization and RLVR-based \citep{lambert2024tulu} optimization procedure in Sec.~\ref{sec:optimization}.

% This explicit generation logic enables execution-based optimization. 
% The method consists of the Text-to-Code-to-TS formulation, synthetic Text-Code-TS triplet construction, and reinforcement learning with execution-based rewards.

\begin{figure}[t]
\centering
\includegraphics[width=\linewidth]{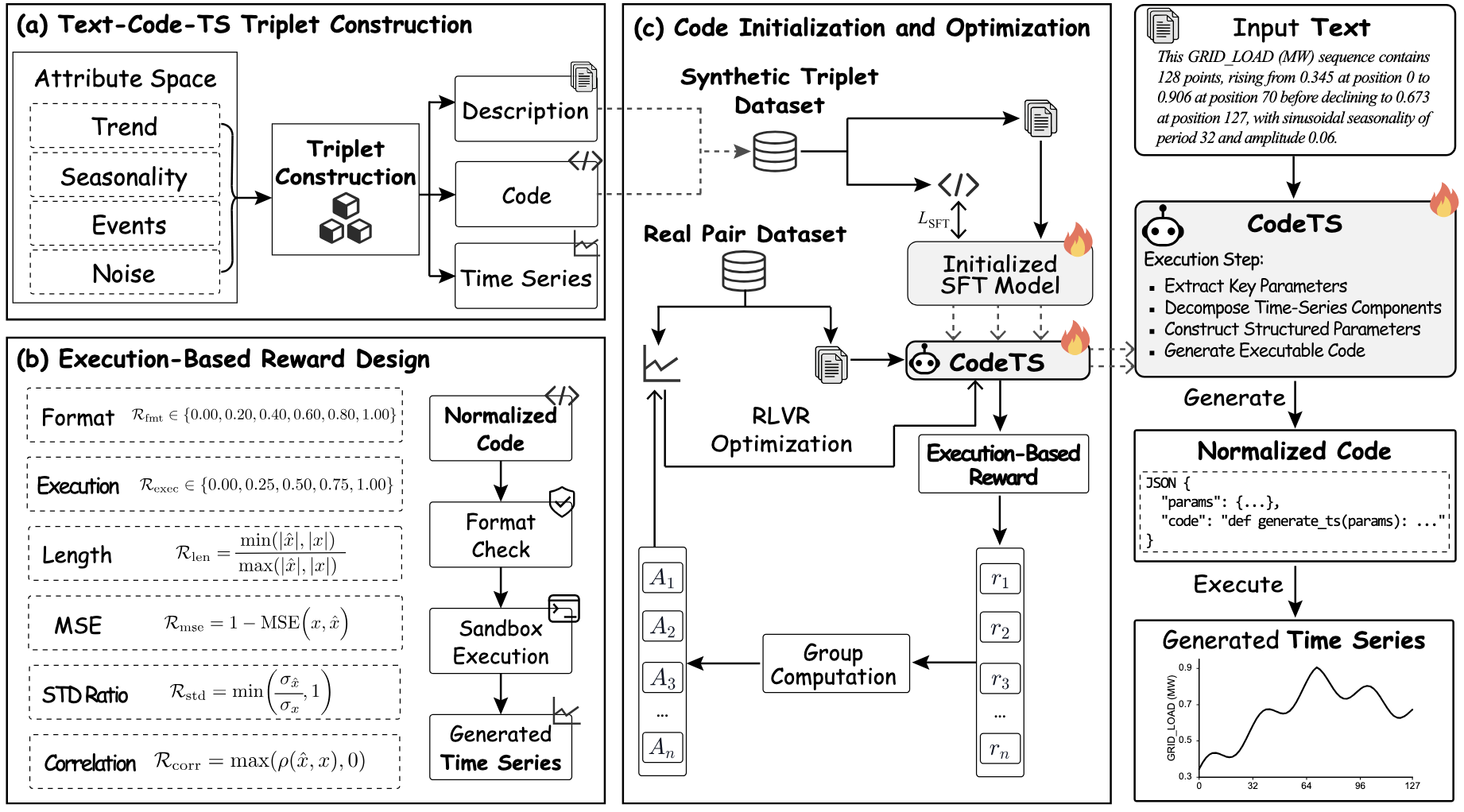}
\caption{Overview of CodeTS. (a) Synthetic Text-Code-TS triplets are constructed from structured temporal attributes. (b) Execution-based rewards verify normalized format, code executability, and time series quality. (c) The model is first initialized on synthetic triplets by SFT and then optimized on real Text-TS pairs with execution-based rewards and GRPO. At inference time, CodeTS maps input text to normalized code and executes it to generate the target time series.}
\label{fig:overview}
\vspace{-0.5em}
\end{figure}

\subsection{Text-to-Code-to-TS Formulation}
\label{sec:formulation}

Conventional Text-to-TS generation directly maps a natural language description \(d\) to a numerical sequence \(x\), leaving the mapping implicit and making it difficult to inspect whether key temporal factors are reflected in the output.
Executable code addresses this limitation by externalizing textual temporal requirements into executable generation programs, making the Text-to-TS mapping explicit, verifiable through execution, and optimizable with execution-based rewards. To this end, we formulate Text-to-TS generation as a Text-to-Code-to-TS process (Fig.~\ref{fig:illustration}(b)). 
Given a natural language description \(d\), the model generates a code program \(y\) from the code space \(\mathcal{Y}\):
% To this end, CodeTS reformulates Text-to-TS generation as a Text-to-Code-to-TS problem (Fig.~\ref{x}). 
% Given a natural language description \(d\), the model first generates executable code \(y \) from the code space \( \mathcal{Y} \):
\begin{equation}
y \sim \pi_\theta(\cdot \mid d),
\label{eq:code_generation}
\end{equation}
where \(\pi_\theta\) denotes the code generation policy. 
The generated time series is then obtained by executing the code:
\begin{equation}
\hat{x} = \mathrm{Execute}(y).
\label{eq:code_execution}
\end{equation}
Therefore, the Text-to-TS generation is formulated as the composition of code generation and execution:
\begin{equation}
d \xrightarrow{\pi_\theta} y \xrightarrow{\mathrm{Execute}} \hat{x}.
\label{eq:text_code_ts_mapping}
\end{equation}
This formulation changes the generation target from raw numerical sequences to executable generation logic, which is well suited to time series because temporal patterns such as trends, periodic fluctuations, local events, and noise can be explicitly specified and combined in code.
However, real world Text-TS pairs lack executable code, which motivates the rule-based construction of synthetic Text-Code-TS triplets in the next subsection.

\subsection{Text-Code-TS Triplet Construction}
\label{sec:triplet}

The Text-to-Code-to-TS formulation requires learning a mapping from textual descriptions to executable code, whereas real world Text-to-TS datasets typically provide only Text-TS pairs without corresponding code annotations.
This motivates the construction of synthetic Text-Code-TS triplets, which provide code-level supervision for initializing the framework. 
% However, directly constructing such triplets from either real time series or free-form text is unreliable, as it may introduce inaccurate descriptions, ambiguous code targets, or misaligned time-series outputs.

Inspired by the synthetic data construction in~\citep{xie2025chatts,lin2026thoth}, we adopt a rule-based triplet construction strategy. 
Each triplet is generated from a shared structured temporal attribute \(a_j\), which specifies the temporal factors to be realized, including sequence length, trend, seasonality, local events, and noise characteristics:
% Specifically, each synthetic sample starts from a structured temporal attribute \(a_j\), which serves as a shared source for generating all elements in the triplet: the textual description, the executable code, and the corresponding time series. 
% The attribute specifies the temporal factors to be realized, including sequence length, trend, seasonality, local events, and noise characteristics:
% Text-to-code-to-TS formulation requires learning a mapping from textual descriptions to executable code, whereas real world Text-to-TS datasets typically provide only Text-TS pairs without corresponding code annotations.
% This motivates the construction of synthetic Text-Code-TS triplets that provide code-level supervision for initializing the framework.
% However, directly constructing Text-Code-TS triplets from either real time series or text is unreliable, as it may introduce inaccurate or misaligned supervision.
% Inspired by the synthetic data construction in~\citep{xie2025chatts,lin2026thoth}, we develop a rule-specified temporal attribute as a shared source to generate aligned descriptions, executable code, and time series. Specifically, each synthetic sample starts from a structured temporal attribute \(a_j\), which serves as the shared source for all elements in the triplet. 
% It specifies the temporal factors to be realized, including sequence length, trend, seasonality, local events, and noise characteristics:
\begin{equation}
a_j =
\left(
a_j^{\mathrm{len}},
a_j^{\mathrm{trend}},
a_j^{\mathrm{season}},
a_j^{\mathrm{event}},
a_j^{\mathrm{noise}}
\right)
\in \mathcal{A},
\label{eq:temporal_attribute}
\end{equation}
where \(\mathcal{A}\) denotes the structured temporal attribute space. 
For synthetic data construction, \(a_j\) serves as a shared source for generating the description, code, and time series, thereby aligning the three elements of each triplet:
\begin{equation}
d_j=\mathcal{G}_{\mathrm{text}}(a_j), 
\qquad
y_j=\mathcal{G}_{\mathrm{code}}(a_j),
\qquad
x_j=\mathrm{Execute}(y_j),
\label{eq:synthetic_triplet_generation}
\end{equation}
where \(\mathcal{G}_{\mathrm{text}}\) and \(\mathcal{G}_{\mathrm{code}}\) generate the textual description and executable code from \(a_j\), respectively. 
The synthetic triplet dataset is then defined as:
\begin{equation}
\mathcal{D}_{\mathrm{syn}}
=
\{(d_j,y_j,x_j)\}_{j=1}^{N_s}.
\label{eq:synthetic_triplet_dataset}
\end{equation}
In each triplet, \(d_j\) is the textual input, \(y_j\) is the reference code for supervised code generation, and \(x_j\) is the execution result of \(y_j\). 
Since all three elements are generated from the same temporal attribute \(a_j\), the triplet is aligned by construction.

\textbf{Normalized Code Representation.} 
{Although the above construction provides aligned descriptions, code, and time series, reliable parsing, execution, and reward computation require a structured code format. We therefore represent each reference code \(y_j\) as a normalized JSON object with two fields: \texttt{params}, which stores temporal parameters such as length, trend, seasonality, local events, and noise settings, and \texttt{code}, which defines a Python function that generates the time series from these parameters. A concrete example is provided in Appendix~\ref{app:normalized_code}.}

\subsection{Execution-Based Reward Design}
\label{sec:reward}

% The normalized code representation allows each generated code to be evaluated through staged verification. 
Given a textual description \(d\), the model samples a code \(y\sim\pi_\theta(\cdot\mid d)\). 
Following Eq.~\eqref{eq:code_execution}, the code is executed to obtain the generated time series \(\hat{x}\) if it passes the required format and execution checks. 
The reward is designed to evaluate both the generated code and its executed output, covering normalized format validity, executable correctness, and time-series quality.

Formally, the overall reward is defined as
\begin{equation}
R(y,x,d)
=
\lambda_{\mathrm{fmt}} R_{\mathrm{fmt}}(y)
+
\lambda_{\mathrm{exec}} R_{\mathrm{exec}}(y)
+
\mathbb{I}_{\mathrm{valid}}(y)\lambda_{\mathrm{ts}} R_{\mathrm{ts}}(\hat{x},x),
\label{eq:overall_reward}
\end{equation}
{where \(R_{\mathrm{fmt}}\) evaluates whether the output follows the normalized \texttt{params}/\texttt{code} schema, \(R_{\mathrm{exec}}\) evaluates whether the generated code is safe and executable, and \(R_{\mathrm{ts}}\), defined in Eq.~\eqref{eq:time_series_reward}, measures the quality of the executed time series. The indicator \(\mathbb{I}_{\mathrm{valid}}(y)\) equals 1 only when the generated output can be correctly parsed and executed; otherwise, the sequence-level reward is not assigned. 
This design prevents invalid code from receiving misleading feedback based on time-series similarity.}
% and \(R_{\mathrm{ts}}\), defined in Eq.~\eqref{eq:time_series_reward}, measures the quality of the executed time series.
% The indicator \(\mathbb{I}_{\mathrm{valid}}(y)\) equals 1 only when the generated output can be correctly parsed and executed; otherwise, the sequence-level reward is not assigned. 
% This design prevents invalid code from receiving misleading feedback based on time-series similarity. 
% \textbf{Format Reward.}
% The format reward checks whether the model output follows the required normalized representation. 
% Specifically, the output should contain both \texttt{params} and \texttt{code} fields, and the \texttt{params} field should include the required time series parameters, such as length, trend, seasonality, local events, and noise-related settings. 
% This reward encourages the model to expose temporal specifications explicitly instead of hiding them inside arbitrary code operations.
% \textbf{Execution Reward.}
% The execution reward evaluates whether the generated code can be safely and successfully executed. 
% Code receives positive execution feedback only if it satisfies the predefined safety constraints, executes without errors, and returns a valid numerical time series. 
% This reward provides code-level feedback before comparing the generated sequence with the target series.
% \textbf{Time-Series Reward.}
After successful execution, the generated time series \(\hat{x}\) is compared with the paired target series \(x\). 
The time series reward is defined as a weighted combination of complementary criteria:
\begin{equation}
R_{\mathrm{ts}}(\hat{x},x)
=
\lambda_{\mathrm{len}}R_{\mathrm{len}}(\hat{x},x)
+
\lambda_{\mathrm{err}}R_{\mathrm{err}}(\hat{x},x)
+
\lambda_{\mathrm{corr}}R_{\mathrm{corr}}(\hat{x},x)
+
\lambda_{\mathrm{stat}}R_{\mathrm{stat}}(\hat{x},x).
\label{eq:time_series_reward}
\end{equation}
Here, \(R_{\mathrm{len}}\) checks length consistency, \(R_{\mathrm{err}}\) measures pointwise numerical accuracy, \(R_{\mathrm{corr}}\) evaluates temporal-pattern alignment through correlation, and \(R_{\mathrm{stat}}\) compares distributional statistics such as scale or variance. 
Together, these terms assess whether the executed code produces a time series that matches the paired target sequence in both local values and global temporal structure. Detailed definitions of these rewards are provided in Appendix~\ref{app:reward_details}.

This reward design turns executable generation into a verifiable optimization signal. For real Text-to-TS pairs where reference code is unavailable, the model can still be optimized through format validation, execution checking, and direct comparison between the generated and target time series.

\subsection{{Initialization and Execution-based Optimization}}
\label{sec:optimization}

The model is trained in two stages. 
First, supervised initialization is performed on the synthetic triplets in Eq.~\eqref{eq:synthetic_triplet_dataset}, where each textual description is paired with a reference executable code. 
The second stage further optimizes the initialized code generation model on real Text-TS pairs using the reward in Eq.~\eqref{eq:overall_reward}, where no reference code is available.

\textbf{Supervised Code Initialization.}
Given the synthetic triplet dataset 
\(\mathcal{D}_{\mathrm{syn}}=\{(d_j,y_j,x_j)\}_{j=1}^{N_s}\), 
the supervised stage trains the model to generate the reference code \(y_j\) conditioned on the textual description \(d_j\). 
The executed series \(x_j\) is not used as a direct generation target in this stage; instead, it verifies that the reference code is executable and aligned with the temporal attribute used to construct the triplet. 
The supervised fine-tuning objective is defined as the negative log-likelihood of the reference code:
\begin{equation}
\mathcal{L}_{\mathrm{SFT}}(\theta)
=
-
\mathbb{E}_{(d,y,x)\sim\mathcal{D}_{\mathrm{syn}}}
\sum_{t=1}^{|y|}
\log \pi_\theta(y_t \mid y_{<t}, d).
\label{eq:sft_objective}
\end{equation}
This stage teaches the model the normalized \texttt{params}/\texttt{code} format and initializes the mapping from textual temporal descriptions to executable generation code. 
Such initialization is important for the subsequent reinforcement learning stage, where invalid or non-executable code would otherwise lead to sparse and noisy feedback.

\textbf{Execution-Guided Policy Optimization.}
After supervised initialization, the model is further optimized on real Text-TS pairs, where no reference code is available. 
For each description \(d_i\), the policy samples a group of \(G\) candidate code. 
Each candidate is checked, executed when valid, and scored by the execution-based reward in Eq.~\eqref{eq:overall_reward}. 
Let \(R_i^{(k)}\) denote the reward of the \(k\)-th sampled code for the pair \((d_i,x_i)\). 
The relative advantage used by GRPO \citep{shao2024deepseekmath} is computed within the sampled group:
\begin{equation}
\hat{A}_i^{(k)}
=
\frac{
R_i^{(k)}
-
\mathrm{mean}\big(\{R_i^{(j)}\}_{j=1}^{G}\big)
}{
\mathrm{std}\big(\{R_i^{(j)}\}_{j=1}^{G}\big)+\varepsilon
},
\label{eq:grpo_advantage}
\end{equation}
where \(\varepsilon\) is a small constant for numerical stability. 
This group-relative normalization encourages code that outperform other candidates generated for the same description, while suppressing invalid or low-quality code.

The policy is updated with GRPO using the relative advantages. 
For stable optimization, we maximize a clipped surrogate objective with KL regularization \citep{schulman2025trust} toward the supervised initialization policy:
\begin{equation}
\mathcal{J}_{\mathrm{GRPO}}(\theta)
=
\mathbb{E}
\left[
\frac{1}{G}
\sum_{k=1}^{G}
\ell_i^{(k)}(\theta)
-
\beta
D_{\mathrm{KL}}
\big(
\pi_\theta(\cdot\mid d_i)
\|
\pi_{\mathrm{SFT}}(\cdot\mid d_i)
\big)
\right],
\label{eq:grpo_objective}
\end{equation}
where the clipped surrogate term is:
% \begin{equation}
% \ell_i^{(k)}(\theta)
% =
% \min
% \left(
% \rho_i^{(k)}(\theta)\hat{A}_i^{(k)},
% \mathrm{clip}\big(\rho_i^{(k)}(\theta),1-\varepsilon,1+\varepsilon\big)\hat{A}_i^{(k)}
% \right).
% \label{eq:grpo_surrogate}
% \end{equation}
\begin{equation}
\ell_i^{(k)}(\theta) = \frac{1}{T_k}
\sum_{t=1}^{T_k}
\min\left(\rho_{i,t}^{(k)}(\theta)\hat{A}_i^{(k)},\mathrm{clip}
\big(\rho_{i,t}^{(k)}(\theta),1-\varepsilon,1+\varepsilon\big)
\hat{A}_i^{(k)}\right).
\label{eq:grpo_surrogate}
\end{equation}
Here, \(T_k\) is the total length of output code, \(\rho_{i, t}^{(k)}(\theta)\) is the policy ratio between the current policy and the sampling policy,  \(\varepsilon\) is the clipping threshold, and \(\beta\) controls the KL regularization strength. 
The clipped surrogate favors candidates with higher relative advantages while limiting overly large policy updates, and the KL term keeps the policy close to the supervised initialization policy.
% Here, \(\rho_i^{(k)}(\theta)\) is the policy ratio between the current policy and the sampling policy, \(\varepsilon\) is the clipping threshold, and \(\beta\) controls the KL regularization strength. 
% The clipped surrogate favors candidates with higher relative advantages while limiting overly large policy updates, and the KL term keeps the policy close to the supervised initialization policy.

Overall, the training process is summarized as:
\begin{equation}
\theta_{\mathrm{SFT}}
=
\arg\min_{\theta}
\mathcal{L}_{\mathrm{SFT}}(\theta),
\qquad
\theta_{\mathrm{final}}
=
\arg\max_{\theta}
\mathcal{J}_{\mathrm{GRPO}}(\theta;\theta_{\mathrm{SFT}}).
\label{eq:two_stage_training}
\end{equation}
The supervised stage initializes executable code generation, and the execution-guided stage improves generation quality through verifiable feedback from code execution and TS comparison.

\section{Experiments}

% We study four questions. (RQ1) asks how CodeTSGen compares with existing methods. (RQ2) examines the contribution of each training stage. (RQ3) studies the effect of the reward design. (RQ4) asks how sensitive the model is to description templates.

\subsection{Experimental Setup}

\textbf{Datasets.}
We evaluate on eight public benchmarks, including ETTh1, ETTh2, ETTm1, ETTm2~\citep{wu2021autoformer}, Electricity, Exchange, Weather~\citep{zhou2021informer}, and Web~\citep{casado2021web}, with generation lengths of 96, 192, and 336. Details of the text-conditioned benchmark and the real Text-TS data used for RLVR training are provided in Appendix~\ref{app:Data_Sources}.

\textbf{Metrics.}
We report Mean Squared Error (MSE), Dynamic Time Warping distance (DTW)~\citep{sakoe2003dynamic}, and Pearson correlation as general quality metrics.
For LLM-based zero-shot baselines, we additionally report Pass@1 \citep{chen2021evaluating} Success Rate (SR), defined as the fraction of samples whose outputs can be successfully parsed and, when applicable, executed. Detailed introductions are provided in Appendix~\ref{app:metrics}.

\textbf{Baselines.}
We compare CodeTS with four supervised generative baselines, including Diffusion-TS~\citep{yuan2024diffusionts}, T2S~\citep{ge2025t2s}, VerbalTS~\citep{gu2025verbalts}, and TimeVQVAE~\citep{lee2023vector}; Diffusion-TS and TimeVQVAE use text-conditioning adapters following~\citet{ge2025t2s}. To adapt non-text generative baselines to the Text-to-TS setting, Diffusion-TS and TimeVQVAE are equipped with text-conditioning adapters~\citep{ge2025t2s}. 
For LLM-based baselines, we evaluate direct zero-shot generators, including ChatTime~\citep{wang2025chattime}, TimeOmni-1~\citep{guan2025timeomni}, and GPT-4o-mini~\citep{hurst2024gpt}, as well as recent coding-capable LLMs under the same Text-to-Code-to-TS prompting setup, including IQuest-Coder-V1-7B-Instruct~\citep{yang2026iquest}, 
Devstral-Small-2-24B-Instruct-2512 (Devstral-Instruct-24B)~\citep{rastogi2025devstral}, Seed-Coder-8B-Instruct~\citep{seed2025seed}, and Qwen3.5-9B~\citep{team2026qwen3}.

\begin{table*}[t]
\vspace{-0.3cm}
\setlength{\belowcaptionskip}{3pt}
\centering
\caption{Comparison with supervised time series generation baselines. CodeTS is evaluated zero-shot, while all supervised baselines are trained separately on each target dataset and generation length. Bold and underline denote the best and second-best results.}
\label{tab:code_vs_supervised}
\Huge
\resizebox{\textwidth}{!}{
\begin{tabular}{ll|ccc|ccc|ccc|ccc|ccc}
\toprule
\cmidrule(lr){3-5}\cmidrule(lr){6-17}
 & & \multicolumn{3}{c|}{CodeTS} & \multicolumn{3}{c|}{VerbalTS} & \multicolumn{3}{c|}{T2S} & \multicolumn{3}{c|}{Diffusion-TS} & \multicolumn{3}{c}{TimeVQVAE} \\
Datasets & Length & MSE$\downarrow$ & DTW$\downarrow$ & Pearson$\uparrow$ & MSE$\downarrow$ & DTW$\downarrow$ & Pearson$\uparrow$ & MSE$\downarrow$ & DTW$\downarrow$ & Pearson$\uparrow$ & MSE$\downarrow$ & DTW$\downarrow$ & Pearson$\uparrow$ & MSE$\downarrow$ & DTW$\downarrow$ & Pearson$\uparrow$ \\
\midrule
\multirow{3}{*}{ETTh1} & 96 & \textbf{0.0202} & \textbf{0.0799} & \textbf{0.8136} & 0.1438 & 0.1794 & 0.0600 & 0.1114 & 0.1740 & -0.0253 & 0.1286 & 0.1721 & 0.1351 & \underline{0.0575} & \underline{0.1545} & \underline{0.3571}\\
 & 192 & \textbf{0.0219} & \textbf{0.0808} & \textbf{0.7714} & 0.1430 & 0.1847 & 0.0889 & 0.1242 & 0.1626 & -0.0331 & 0.0933 & \underline{0.1415} & 0.1620 & \underline{0.0578} & 0.1533 & \underline{0.3005}\\
 & 336 & \textbf{0.0232} & \textbf{0.0823} & \textbf{0.7242} & 0.1895 & 0.2368 & 0.0414 & 0.0839 & 0.1522 & -0.0474 & 0.0835 & \underline{0.1385} & \underline{0.2414} & \underline{0.0666} & 0.2061 & 0.0608 \\
\midrule
\multirow{3}{*}{ETTh2} & 96 & \textbf{0.0305} & \textbf{0.1041} & \textbf{0.7302} & 0.1238 & 0.1891 & \underline{0.2522} & 0.1126 & 0.1931 & 0.0150 & 0.1259 & 0.2118 & 0.2064 & \underline{0.0741} & \underline{0.1769} & 0.2389 \\
 & 192 & \textbf{0.0295} & \textbf{0.0999} & \textbf{0.6835} & 0.1598 & 0.2220 & 0.0668 & 0.1277 & 0.2078 & 0.0199 & 0.1465 & 0.2312 & 0.1246 & \underline{0.0852} & \underline{0.1977} & \underline{0.2234}\\
 & 336 & \textbf{0.0292} & \textbf{0.0993} & \textbf{0.6456} & 0.2531 & 0.3584 & 0.0336 & 0.1003 & 0.1945 & 0.0251 & 0.1059 & \underline{0.1868} & \underline{0.1001} & \underline{0.0935} & 0.2368 & 0.0202 \\
\midrule
\multirow{3}{*}{ETTm1} & 96 & \textbf{0.0161} & \textbf{0.0669} & \textbf{0.8713} & \underline{0.0723} & \underline{0.1139} & \underline{0.4928} & 0.1143 & 0.1998 & 0.0069 & 0.1480 & 0.1824 & 0.0386 & 0.0830 & 0.2338 & 0.0434 \\
 & 192 & \textbf{0.0181} & \textbf{0.0685} & \textbf{0.8338} & 0.0854 & \underline{0.1274} & \underline{0.3767} & 0.1104 & 0.1900 & -0.0123 & 0.1246 & 0.1595 & 0.1136 & \underline{0.0679} & 0.1647 & 0.3147 \\
 & 336 & \textbf{0.0276} & \textbf{0.0704} & \textbf{0.7341} & 0.1429 & 0.1786 & 0.0011 & 0.0730 & 0.1564 & 0.0257 & 0.1124 & \underline{0.1450} & \underline{0.0264} & \underline{0.0619} & 0.1933 & 0.0000 \\
\midrule
\multirow{3}{*}{ETTm2} & 96 & \textbf{0.0279} & \textbf{0.0982} & \textbf{0.7944} & 0.0883 & \underline{0.1354} & \underline{0.4486} & 0.1604 & 0.2525 & 0.0125 & 0.1820 & 0.2372 & 0.0266 & \underline{0.0882} & 0.2253 & 0.0629 \\
 & 192 & \textbf{0.0295} & \textbf{0.1016} & \textbf{0.7532} & 0.1091 & 0.1608 & 0.2999 & 0.1167 & 0.1925 & 0.0298 & 0.1434 & 0.1994 & 0.1579 & \underline{0.0705} & \underline{0.1556} & \underline{0.3092}\\
 & 336 & \textbf{0.0332} & \textbf{0.0968} & \textbf{0.6689} & 0.1305 & \underline{0.1830} & -0.0058 & 0.0945 & 0.1838 & \underline{0.0331} & 0.1320 & 0.1940 & 0.0118 & \underline{0.0671} & 0.2117 & -0.0111 \\
\midrule
\multirow{3}{*}{Weather} & 96 & \textbf{0.0159} & \textbf{0.0622} & \textbf{0.8599} & \underline{0.0741} & \underline{0.0874} & \underline{0.4925} & 0.1636 & 0.2468 & 0.0023 & 0.1566 & 0.1839 & 0.0374 & 0.0907 & 0.2312 & 0.0108 \\
 & 192 & \textbf{0.0159} & \textbf{0.0592} & \textbf{0.8427} & \underline{0.0787} & \underline{0.0930} & \underline{0.4415} & 0.1652 & 0.2370 & 0.0284 & 0.1566 & 0.1894 & 0.0188 & 0.0812 & 0.2114 & 0.0044 \\
 & 336 & \textbf{0.0213} & \textbf{0.0648} & \textbf{0.7701} & 0.1344 & \underline{0.1409} & -0.0026 & 0.1105 & 0.1951 & \underline{0.0219} & 0.1532 & 0.1986 & 0.0097 & \underline{0.0721} & 0.2008 & -0.0045 \\
\midrule
\multirow{3}{*}{Electricity} & 96 & \textbf{0.0301} & \textbf{0.0923} & \textbf{0.8112} & \underline{0.0458} & \underline{0.0968} & \underline{0.7666} & 0.1725 & 0.2206 & 0.0629 & 0.1033 & 0.1389 & 0.5250 & 0.0559 & 0.1027 & 0.7242 \\
 & 192 & \textbf{0.0284} & \textbf{0.0845} & \textbf{0.7982} & \underline{0.0461} & \underline{0.1024} & \underline{0.7372} & 0.1515 & 0.1869 & 0.1028 & 0.0966 & 0.1421 & 0.5896 & 0.0657 & 0.1341 & 0.6669 \\
 & 336 & \textbf{0.0280} & \textbf{0.0824} & \textbf{0.7811} & \underline{0.0533} & \underline{0.1197} & \underline{0.7024} & 0.1121 & 0.1550 & 0.1137 & 0.1020 & 0.1585 & 0.5294 & 0.0670 & 0.1620 & 0.6459 \\
\midrule
\multirow{3}{*}{Exchange} & 96 & \textbf{0.0795} & \textbf{0.1643} & \textbf{0.4815} & 0.2182 & 0.2499 & 0.0318 & \underline{0.1296} & \underline{0.2235} & 0.0013 & 0.2110 & 0.2906 & 0.0516 & 0.1660 & 0.3090 & \underline{0.2021}\\
 & 192 & \textbf{0.0837} & \textbf{0.1481} & \textbf{0.5180} & 0.2714 & 0.2979 & 0.0083 & 0.1777 & \underline{0.2634} & 0.0059 & 0.2064 & 0.2794 & \underline{0.1936} & \underline{0.1038} & 0.2646 & 0.0447 \\
 & 336 & \textbf{0.0870} & \textbf{0.1449} & \textbf{0.5159} & 0.2336 & 0.3207 & 0.0047 & 0.1090 & \underline{0.1921} & -0.0246 & 0.1315 & 0.2021 & \underline{0.1139} & \underline{0.0957} & 0.2508 & 0.0040 \\
\midrule
\multirow{3}{*}{Web} & 96 & \textbf{0.0326} & \underline{0.1162} & \textbf{0.3672} & 0.0518 & \textbf{0.0940} & \underline{0.0533} & 0.0727 & 0.1597 & -0.0147 & 0.0582 & 0.1169 & -0.0123 & \underline{0.0430} & 0.1274 & -0.0411 \\
 & 192 & \textbf{0.0246} & 0.1002 & \textbf{0.3024} & 0.0343 & \textbf{0.0733} & \underline{0.0577} & 0.0391 & 0.1056 & -0.0147 & 0.0360 & \underline{0.0909} & -0.0080 & \underline{0.0277} & 0.0925 & -0.0270 \\
 & 336 & \underline{0.0175} & 0.0823 & \textbf{0.2664} & 0.0273 & 0.0751 & 0.0203 & 0.0212 & 0.0749 & 0.0209 & 0.0219 & \underline{0.0662} & 0.0463 & \textbf{0.0171} & \textbf{0.0653} & \underline{0.1078}\\
\midrule
\multicolumn{2}{c|}{Average} & \textbf{0.0321} & \textbf{0.0938} & \textbf{0.6808} & 0.1213 & \underline{0.1675} & \underline{0.2279} & 0.1148 & 0.1883 & 0.0148 & 0.1233 & 0.1774 & 0.1433 & \underline{0.0733} & 0.1859 & 0.1774 \\
\bottomrule
\end{tabular}
}
\vspace{-0.2cm}
\end{table*}

\begin{table*}[t]
\vspace{-0.2cm}
\setlength{\belowcaptionskip}{3pt}
\centering
\caption{Comparison with zero-shot LLM baselines for time series generation. SR denotes the pass@1 execution success rate. $^{\dagger}$ indicates values that are approximately 100.0\%. Bold and underline denote the best and second-best results.}
\label{tab:zero_shot_sr}
\Huge
\resizebox{\textwidth}{!}{
\begin{tabular}{ll|cccc|cccc|cccc|cccc}
\toprule
\cmidrule(lr){3-18}
 & & \multicolumn{4}{c|}{CodeTS} & \multicolumn{4}{c|}{TimeOmni-1} & \multicolumn{4}{c|}{ChatTime} & \multicolumn{4}{c}{GPT4o-mini} \\
Datasets & Length & MSE$\downarrow$ & DTW$\downarrow$ & Pearson$\uparrow$ & SR(\%)$\uparrow$ & MSE$\downarrow$ & DTW$\downarrow$ & Pearson$\uparrow$ & SR(\%)$\uparrow$ & MSE$\downarrow$ & DTW$\downarrow$ & Pearson$\uparrow$ & SR(\%)$\uparrow$ & MSE$\downarrow$ & DTW$\downarrow$ & Pearson$\uparrow$ & SR(\%)$\uparrow$ \\
\midrule
\multirow{3}{*}{ETTh1} & 96 & \textbf{0.0202} & \textbf{0.0799} & \textbf{0.8136} & {100.0} & 0.1749 & \underline{0.2736} & 0.0669 & 87.1 & 0.1826 & 0.2772 & 0.0063 & 92.0 & \underline{0.1721} & 0.3344 & \underline{0.0810} & \underline{97.3}\\
 & 192 & \textbf{0.0219} & \textbf{0.0808} & \textbf{0.7714} & {100.0} & 0.1592 & 0.2644 & \underline{0.0660} & 82.9 & 0.1672 & \underline{0.2561} & 0.0156 & 92.9 & \underline{0.1552} & 0.3201 & 0.0527 & \underline{93.7}\\
 & 336 & \textbf{0.0232} & \textbf{0.0823} & \textbf{0.7242} & {100.0} & 0.1615 & 0.2737 & \underline{0.0499} & 66.0 & 0.1536 & \underline{0.2458} & 0.0182 & \underline{89.8} & \underline{0.1488} & 0.3201 & 0.0098 & 87.1\\
\midrule
\multirow{3}{*}{ETTh2} & 96 & \textbf{0.0305} & \textbf{0.1041} & \textbf{0.7302} & {100.0} & 0.2489 & 0.3131 & \underline{0.0368} & 79.3 & \underline{0.1771} & \underline{0.2814} & -0.0020 & 90.8 & 0.3353 & 0.5147 & 0.0152 & \underline{97.1}\\
 & 192 & \textbf{0.0295} & \textbf{0.0999} & \textbf{0.6835} & {100.0} & 0.2094 & 0.3091 & \underline{0.0514} & 77.8 & \underline{0.1730} & \underline{0.2740} & -0.0087 & 88.9 & 0.3237 & 0.5116 & 0.0082 & \underline{92.1}\\
 & 336 & \textbf{0.0292} & \textbf{0.0993} & \textbf{0.6456} & {100.0} & 0.2047 & 0.3078 & \underline{0.0441} & 56.5 & \underline{0.1450} & \underline{0.2432} & -0.0446 & \underline{91.2} & 0.2995 & 0.4919 & -0.0048 & 74.1\\
\midrule
\multirow{3}{*}{ETTm1} & 96 & \textbf{0.0161} & \textbf{0.0669} & \textbf{0.8713} & {100.0} & 0.1967 & \underline{0.2755} & 0.1279 & 86.4 & \underline{0.1924} & 0.2856 & -0.0050 & 92.7 & 0.2004 & 0.3543 & \underline{0.1668} & \underline{97.3}\\
 & 192 & \textbf{0.0181} & \textbf{0.0685} & \textbf{0.8338} & {100.0} & 0.1836 & 0.2759 & \underline{0.0796} & 80.5 & \underline{0.1742} & \underline{0.2651} & 0.0011 & 91.6 & 0.1808 & 0.3419 & 0.0772 & \underline{94.3}\\
 & 336 & \textbf{0.0276} & \textbf{0.0704} & \textbf{0.7341} & {100.0} & 0.1669 & 0.2695 & \underline{0.0548} & 66.1 & \underline{0.1607} & \underline{0.2538} & -0.0039 & 86.4 & 0.1702 & 0.3417 & 0.0136 & \underline{91.6}\\
\midrule
\multirow{3}{*}{ETTm2} & 96 & \textbf{0.0279} & \textbf{0.0982} & \textbf{0.7944} & {100.0} & 0.2618 & 0.3079 & \underline{0.0901} & 80.2 & \underline{0.1983} & \underline{0.2955} & -0.0038 & 92.5 & 0.3669 & 0.5284 & 0.0134 & \underline{96.8}\\
 & 192 & \textbf{0.0295} & \textbf{0.1016} & \textbf{0.7532} & {100.0} & 0.2423 & 0.3209 & \underline{0.0484} & 73.4 & \underline{0.1815} & \underline{0.2733} & 0.0001 & 91.3 & 0.3500 & 0.5255 & 0.0009 & \underline{93.1}\\
 & 336 & \textbf{0.0332} & \textbf{0.0968} & \textbf{0.6689} & {100.0} & 0.2242 & 0.3200 & \underline{0.0521} & 59.7 & \underline{0.1578} & \underline{0.2538} & -0.0016 & \underline{86.7} & 0.3382 & 0.5223 & -0.0005 & 84.5\\
\midrule
\multirow{3}{*}{Weather} & 96 & \textbf{0.0159} & \textbf{0.0622} & \textbf{0.8599} & {100.0} & \underline{0.2330} & \underline{0.2748} & \underline{0.1505} & 79.1 & 0.2400 & 0.3338 & -0.0040 & 92.9 & 0.3276 & 0.4507 & 0.0758 & \underline{98.0}\\
 & 192 & \textbf{0.0159} & \textbf{0.0592} & \textbf{0.8427} & {100.0} & 0.2586 & \underline{0.3068} & \underline{0.0680} & 68.1 & \underline{0.2375} & 0.3273 & 0.0010 & 91.5 & 0.2981 & 0.4279 & 0.0341 & \underline{95.0}\\
 & 336 & \textbf{0.0213} & \textbf{0.0648} & \textbf{0.7701} & {100.0} & \underline{0.2144} & \underline{0.2765} & \underline{0.0858} & 59.2 & 0.2206 & 0.3112 & 0.0102 & 87.2 & 0.2512 & 0.3928 & 0.0234 & \underline{88.1}\\
\midrule
\multirow{3}{*}{Electricity} & 96 & \textbf{0.0301} & \textbf{0.0923} & \textbf{0.8112} & {100.0} & 0.2490 & \underline{0.2817} & \underline{0.0123} & 79.3 & \underline{0.2165} & 0.2903 & 0.0045 & 92.8 & 0.3337 & 0.4789 & 0.0054 & \underline{97.0}\\
 & 192 & \textbf{0.0284} & \textbf{0.0845} & \textbf{0.7982} & {99.8} & 0.2186 & 0.2998 & \underline{0.0034} & 73.6 & \underline{0.2106} & \underline{0.2847} & 0.0006 & 90.1 & 0.3161 & 0.4699 & 0.0024 & \underline{94.9}\\
 & 336 & \textbf{0.0280} & \textbf{0.0824} & \textbf{0.7811} & {99.9} & 0.1887 & 0.2992 & \underline{0.0205} & 65.0 & \underline{0.1858} & \underline{0.2711} & 0.0000 & 84.9 & 0.3000 & 0.4611 & 0.0036 & \underline{85.3}\\
\midrule
\multirow{3}{*}{Exchange} & 96 & \textbf{0.0795} & \textbf{0.1643} & \textbf{0.4815} & {100.0} & 0.3052 & 0.3530 & \underline{0.0577} & 64.3 & \underline{0.2118} & \underline{0.3079} & 0.0411 & 96.2 & 0.4009 & 0.4932 & 0.0264 & \underline{97.3}\\
 & 192 & \textbf{0.0837} & \textbf{0.1481} & \textbf{0.5180} & {100.0} & 0.2893 & 0.3536 & 0.0324 & 51.6 & \underline{0.1924} & \underline{0.2956} & 0.0213 & \underline{92.3} & 0.3706 & 0.4718 & \underline{0.0619} & \underline{92.3}\\
 & 336 & \textbf{0.0870} & \textbf{0.1449} & \textbf{0.5159} & {100.0} & 0.2927 & 0.3896 & \underline{0.0353} & 28.6 & \underline{0.1983} & \underline{0.3133} & -0.0800 & \underline{91.1} & 0.3415 & 0.4784 & 0.0301 & 71.4\\
\midrule
\multirow{3}{*}{Web} & 96 & \textbf{0.0326} & \textbf{0.1162} & \textbf{0.3672} & {100.0} & 0.1738 & 0.2314 & \underline{0.0639} & 81.6 & 0.3097 & 0.4230 & -0.0012 & \underline{92.0} & \underline{0.0631} & \underline{0.1690} & 0.0577 & 90.0\\
 & 192 & \textbf{0.0246} & \textbf{0.1002} & \textbf{0.3024} & {100.0} & 0.2420 & 0.2924 & 0.0476 & 65.0 & 0.3359 & 0.4486 & 0.0001 & \underline{91.8} & \underline{0.0406} & \underline{0.1293} & \underline{0.0509} & 67.7\\
 & 336 & \textbf{0.0175} & \textbf{0.0823} & \textbf{0.2664} & {99.9} & 0.2387 & 0.2776 & 0.0242 & 48.9 & 0.3313 & 0.4483 & -0.0099 & \underline{87.5} & \underline{0.0254} & \underline{0.0997} & \underline{0.0657} & 37.3\\
\midrule
\multicolumn{2}{c|}{Average} & \textbf{0.0321} & \textbf{0.0938} & \textbf{0.6808} & {100.0$^{\dagger}$ } & 0.2224 & \underline{0.2978} & \underline{0.0571} & 69.2 & \underline{0.2064} & 0.3025 & -0.0019 & \underline{90.7} & 0.2546 & 0.4012 & 0.0363 & 88.1 \\
\bottomrule
\end{tabular}
}
\vspace{-0.5cm}
\end{table*}

\textbf{Implementation Details.}
CodeTS is initialized from Qwen2.5-Coder-7B-Instruct~\citep{hui2024qwen25coder}. 
For SFT warmup, we train on 500 synthetic triplets for one epoch with a maximum sequence length of 3072, a learning rate of $5\times10^{-6}$, a batch size of 64.
For RLVR, we train on 6300 real $(d,x)$ pairs using GRPO with 4 samples per prompt, learning rate $5\times10^{-7}$, KL coefficient $10^{-3}$, and 200 update steps. 
We use ZeRO-2~\citep{rajbhandari2020zero} for SFT and ZeRO-3 with vLLM~\citep{kwon2023efficient} for RLVR on A100 GPUs. More details are provided in Appendix~\ref{app:training_config}.

% CodeTS uses Qwen2.5-Coder-7B-Instruct~\citep{hui2024qwen25coder}.
% The SFT warmup stage uses 500 synthetic triplets with a maximum sequence length of 3072, a learning rate of $5\times10^{-6}$, a batch size of 64, and one training epoch.
% The RLVR stage uses 6300 real $(d, x)$ pairs and applies GRPO with 4 samples per prompt, a learning rate of $5\times10^{-7}$, a KL coefficient of $10^{-3}$, a clipping range of $[\varepsilon_{\text{low}}, \varepsilon_{\text{high}}]=[0.2, 0.28]$, and 200 update steps.
% The reward weights are $\lambda_{\mathrm{fmt}} = 0.05$, $\lambda_{\mathrm{exec}} = 0.10$, $\lambda_{\mathrm{len}} = 0.10$, $\lambda_{\mathrm{err}} = 0.30$, $\lambda_{\mathrm{corr}} = 0.30$, and $\lambda_{\mathrm{stat}} = 0.15$.
% We use ZeRO-2~\citep{rajbhandari2020zero} for SFT warmup and ZeRO-3 with vLLM~\citep{kwon2023efficient} for RLVR on A100 GPUs.

\subsection{Main Results}

In Table~\ref{tab:code_vs_supervised}, CodeTS achieves the best averaged performance across all three metrics, despite being evaluated zero-shot against supervised baselines trained separately on each target dataset and generation length. Compared with the strongest supervised baseline for each metric, CodeTS improves MSE from 0.0733 to 0.0321, DTW from 0.1675 to 0.0938, and Pearson correlation from 0.2279 to 0.6808, corresponding to relative improvements of~\textbf{56.2\%, 44.0\%, and 198.7\%}, respectively. On the Web dataset, although CodeTS does not always achieve the best MSE or DTW, it consistently obtains the best Pearson correlation across all generation lengths. This metric is particularly informative in this dataset, where long flat regions and sparse abrupt variations make shape preservation more relevant than pointwise error alone. We provide a detailed analysis of the Web results in {Appendix~\ref{app:web_analysis}}.
Table~\ref{tab:zero_shot_sr} shows that CodeTS achieves the best performance across all length settings and all reported metrics. 
In the averaged results, CodeTS reduces MSE from 0.2064 to 0.0321 and DTW from 0.2978 to 0.0938 compared with the strongest zero-shot LLM baseline for each error metric, corresponding to relative reductions of ~\textbf{84.4\%} and~\textbf{68.5\%}, respectively. 
It also increases the average Pearson correlation from 0.0571 to 0.6808, corresponding to \textbf{11.9} times the strongest zero-shot LLM baseline, while maintaining a near-perfect pass@1 execution success rate. 

We further compare with mainstream and recent strong coding-capable LLMs, including IQuest-Coder-V1-7B-Instruct~\citep{yang2026iquest}, Devstral-Instruct-24B~\citep{rastogi2025devstral}, Seed-Coder-8B-Instruct~\citep{seed2025seed}, and Qwen3.5-9B~\citep{team2026qwen3}, under the same Text-to-Code-to-TS setup. Although these models achieve high execution success rates, CodeTS remains stronger overall, especially in Pearson correlation, indicating better preservation of temporal patterns rather than merely producing executable code. Detailed comparisons are provided in Appendix~\ref{app:code_llm_baselines}.

% \ckx{To further rule out the possibility that the gains come only from using a code-capable backbone, we also compare with several comparable-scale code LLMs under the same Text-to-Code-to-TS setup in Appendix  Table~\ref{tab:comparable_llm_scale}; CodeTS remains consistently stronger in both generation quality and execution reliability.}

\subsection{Ablation Studies}

We conduct ablation studies on three key components of CodeTS: the training stages, the normalized code representation, and the reward design; the main paper reports averaged and representative results, with detailed per-dataset and per-length statistics provided in Appendix~\ref{app:ablation_details}. 
Additional supporting analyses, including RLVR training dynamics, prompt settings, and representative case studies, are provided in Appendix~\ref{app:rl_dynamics}--\ref{app:case_studies}.

% We conduct ablation studies on three key components of CodeTS: the training stages, the normalized code representation, and the reward design. 
% Additional per-dataset and per-length results, together with RLVR training dynamics, prompt settings, and representative case studies, are provided in Appendix~\ref{app:ablation_details}--\ref{app:case_studies}.

\begin{figure}
\vspace{-0.4em}
\centering
\setlength{\abovecaptionskip}{2pt}
\setlength{\belowcaptionskip}{0pt}
\begin{minipage}[t]{0.38\textwidth}
\vspace{0pt}
\centering
\captionof{table}{Training-stage ablation averaged over all benchmark datasets and output lengths.}
\label{tab:training_stage_ablation_overall}
\vspace{0.25em}
\fontsize{7}{6.8}\selectfont
\setlength{\tabcolsep}{7.6pt}
\renewcommand{\arraystretch}{1.15}
\begin{tabular}{@{}ccccc@{}}
\toprule
\textbf{Length} & \textbf{Metric} & \textbf{Base} & \textbf{SFT} & \textbf{Full} \\
\midrule
\multirow{3}{*}{\textbf{96}} & $\downarrow$MSE & 0.095 & 0.089 & \textbf{0.032} \\
 & $\downarrow$DTW & 0.147 & 0.141 & \textbf{0.098} \\
 & $\uparrow$PC & 0.238 & 0.282 & \textbf{0.716} \\
\midrule
\multirow{3}{*}{\textbf{192}} & $\downarrow$MSE & 0.092 & 0.086 & \textbf{0.031} \\
 & $\downarrow$DTW & 0.137 & 0.134 & \textbf{0.093} \\
 & $\uparrow$PC & 0.185 & 0.210 & \textbf{0.688} \\
\midrule
\multirow{3}{*}{\textbf{336}} & $\downarrow$MSE & 0.084 & 0.081 & \textbf{0.033} \\
 & $\downarrow$DTW & 0.135 & 0.133 & \textbf{0.090} \\
 & $\uparrow$PC & 0.132 & 0.157 & \textbf{0.638} \\
\bottomrule
\end{tabular}
\end{minipage}\hfill
\begin{minipage}[t]{0.60\textwidth}
\vspace{0pt}
\centering
\includegraphics[width=\linewidth]{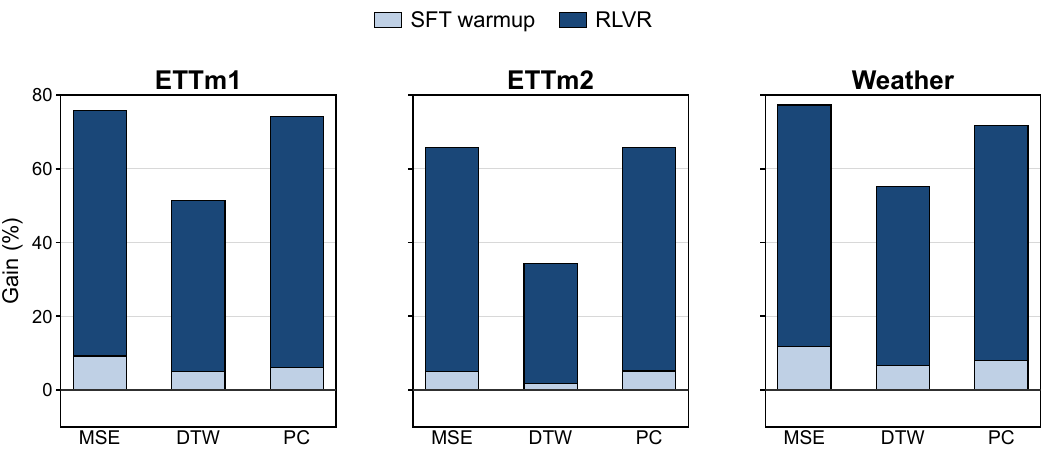}
\vspace{-1.30em}
\captionof{figure}{Stage-wise gains from SFT and RLVR on three datasets. Stacked bars show the percentage improvement contributed by each stage.}
\label{fig:training_stage_ablation}
\end{minipage}
\vspace{-0.0em}
\end{figure}

\textbf{Training Stage Ablation.}
We analyze the contribution of each training stage by comparing the base model, the SFT-initialized model, and the final RLVR-optimized model. 
As shown in Table~\ref{tab:training_stage_ablation_overall}, the averaged performance over all benchmark datasets improves from the base model to SFT and further to the full RLVR-optimized model across output lengths. 
This shows that both stages are beneficial: SFT improves MSE, DTW, and Pearson correlation by learning the normalized code format and basic executable generation patterns, while RLVR further improves the executed time series through reward-based optimization on real Text-TS pairs. 
The improvement from SFT is moderate because it mainly serves as a warmup for format alignment and executable code initialization, rather than directly optimizing the generated series against real targets. 
Fig.~\ref{fig:training_stage_ablation} further decomposes the relative gains on representative datasets, showing that SFT provides a stable warmup contribution, whereas RLVR accounts for the dominant share of the improvement, especially on Pearson correlation. 

% \textbf{Training Stage Ablation.}
% We analyze the contribution of each training stage by comparing the base model, the SFT-initialized model, and the final RLVR-optimized model. 
% As shown in Table~\ref{tab:training_stage_ablation_overall}, the averaged performance over all benchmark datasets improves from the base model to the SFT-initialized model and further to the full RLVR-optimized model across output lengths. This indicates that both stages are beneficial: SFT improves MSE, DTW, and Pearson correlation by learning the normalized code format and basic executable generation patterns, while RLVR further improves the executed time series through reward-based optimization on real Text-TS pairs. The improvement from SFT is relatively moderate because it mainly serves as a warmup for format alignment and executable code initialization, rather than directly optimizing the generated series against real targets. Fig.~\ref{fig:training_stage_ablation} further decomposes the relative gains on representative datasets. The stacked bars show that SFT provides a stable warmup contribution, whereas RLVR accounts for the dominant share of the overall improvement, especially on Pearson correlation. These results suggest that SFT prepares the model for executable code generation, while RLVR is the key stage for aligning generated series with real temporal structures. 

\begin{figure}[!htbp]
\vspace{-0.0em}
\centering
\setlength{\abovecaptionskip}{2pt}
\setlength{\belowcaptionskip}{0pt}
\begin{minipage}[t]{0.49\textwidth}
\vspace{0pt}
\centering
\includegraphics[width=\linewidth]{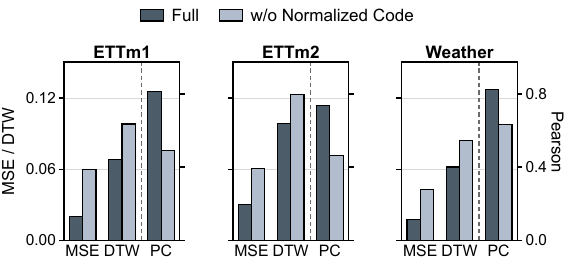}
\vspace{-0em}
\centerline{\small\textbf{(a)} Normalized code}
\end{minipage}\hfill
\begin{minipage}[t]{0.49\textwidth}
\vspace{0pt}
\centering
\includegraphics[width=\linewidth]{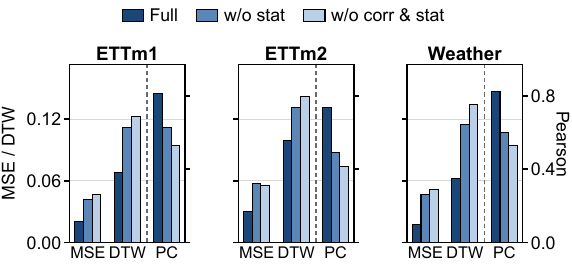}
\vspace{-0em}
\centerline{\small\textbf{(b)} Reward function}
\end{minipage}
\caption{Ablation results for normalized code representation and reward design. 
Panel (a) studies normalized code, and panel (b) studies reward components.}
\label{fig:ablation_panels}
\vspace{-0.6em}
\end{figure}

\textbf{Normalized Code Representation Ablation.}
We evaluate the effect of the normalized code representation by comparing full CodeTS with a variant that directly generates executable code without explicit temporal parameters. As shown in Fig.~\ref{fig:ablation_panels}(a), removing normalized code consistently worsens the results on ETTm1, ETTm2, and Weather, leading to higher MSE and DTW and lower Pearson correlation. This shows that the improvement of CodeTS does not come only from using executable code, but also from structuring the output into explicit temporal parameters and executable logic. 

\textbf{Reward Design Ablation.}
We evaluate the reward design by removing key components from the time-series quality reward while keeping the format and execution rewards fixed. As shown in Fig.~\ref{fig:ablation_panels}(b), the full reward achieves the best overall performance across the evaluated datasets, with lower MSE and DTW and higher Pearson correlation. Removing the statistical reward weakens the results, and removing both correlation and statistical rewards leads to further degradation, especially in Pearson correlation.

\begin{wrapfigure}{r}{0.50\textwidth}
\vspace{-0.5em}
\centering
\setlength{\abovecaptionskip}{2pt}
\setlength{\belowcaptionskip}{0pt}
\includegraphics[width=\linewidth]{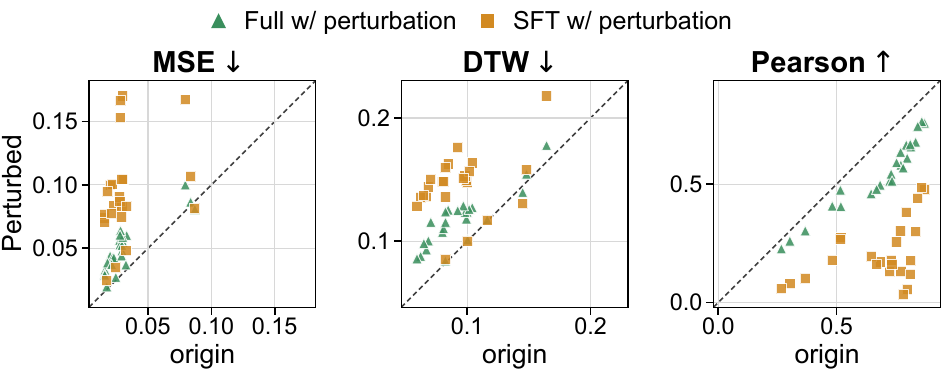}
\caption{Template robustness analysis under perturbed descriptions.}
\label{fig:template_dependency_metric_scatter}
\vspace{-1.0em}
\end{wrapfigure}

\subsection{Template Dependency Analysis}

We further evaluate whether CodeTS remains effective when the input description is written in a different form. Specifically, we rewrite each description by preserving the same temporal attributes while changing the wording and attribute order, and then test the model on these perturbed descriptions. As shown in Figure~\ref{fig:template_dependency_metric_scatter}, performance generally decreases after perturbation, indicating that the model is still affected by changes in description form. However, the final RLVR-optimized model remains much closer to the diagonal line than the SFT-warmup model across MSE, DTW, and Pearson correlation. Since points closer to the diagonal indicate smaller performance changes after perturbation, this result shows that RLVR improves robustness to textual perturbations and helps CodeTS learn a more stable mapping from temporal descriptions to generated time series, rather than relying only on fixed training templates. Detailed analysis are provided in Appendix~\ref{app:template_robustness_details}.

\FloatBarrier
\section{Conclusion}

We presented CodeTS, a verifiable Text-to-Code-to-TS framework that uses executable code as an explicit intermediate interface for Text-to-Time Series Generation. Through Text-Code-TS triplet initialization and execution-based RLVR, CodeTS learns to generate executable programs that translate textual temporal requirements into time series and can be optimized through verifiable feedback. 
Experiments on eight benchmarks show that CodeTS achieves strong Text-to-TS generation performance, outperforming LLM-based baselines and achieving better averaged results than supervised generative baselines trained on the target datasets.

% Experiments on eight benchmarks show that CodeTS achieves strong Text-to-TS generation performance, clearly outperforming LLM-based baselines and achieving better averaged results than supervised generative baselines trained on the target datasets.

% Experiments on eight benchmarks show that CodeTS achieves strong zero-shot Text-to-TS generation performance, clearly outperforming LLM-based zero-shot baselines and achieving better averaged results than supervised generative baselines trained on the target datasets.

% Extensive experiments on eight benchmarks show that CodeTS achieves state-of-the-art zero-shot Text-to-TS performance, consistently outperforming supervised Text-to-TS methods and LLM-based baselines.

\textbf{Limitations and Future Work.}
While CodeTS demonstrates strong zero-shot generation ability, its current formulation is limited to univariate time series and can still be sensitive to ambiguous descriptions or missing temporal attributes. Future work will extend CodeTS to multivariate temporal data, and explore richer execution-based rewards for more data-scarce and high-value application generation scenarios.

% We proposed CodeTS, a verifiable Text-to-Code-to-TS framework that uses executable code as an intermediate interface for Text-to-Time Series Generation. Through Text-Code-TS triplet initialization and execution-based RLVR, CodeTS learns to generate executable code aligned with textual temporal requirements. Experiments on eight benchmarks demonstrate that CodeTS achieves state-of-the-art zero-shot performance, outperforming supervised Text-to-TS methods and LLM-based baselines.

% \textbf{Limitations and Future Work.}
% While CodeTS shows promising performance in zero-shot Text-to-TS generation, its current formulation mainly focuses on univariate time series and may still face difficulties when textual descriptions are ambiguous or lack clear temporal attributes. Future research will extend CodeTS to more complex multivariate temporal settings, improve robustness to ambiguous natural-language descriptions, and develop richer execution-based rewards for realistic time series generation scenarios.

\bibliographystyle{plainnat}
\bibliography{references}

\newpage
\appendix
\section{Notations}
\label{app:notations}

Table~\ref{tab:notations} summarizes the main mathematical symbols used in the
paper.

\begin{center}
\centering
\captionof{table}{Notations used in CodeTS.}
\label{tab:notations}
\setlength{\tabcolsep}{5pt}
\renewcommand{\arraystretch}{1.12}
\begin{tabular}{p{0.39\linewidth}p{0.53\linewidth}}
\toprule
Notation & Description \\
\midrule
\(\mathcal{D}=\{(x_i,d_i)\}_{i=1}^{N}\) & Text-TS dataset. \\
\(d_i\), \(d\) & Text description. \\
\(x_i\), \(x\) & Target time series. \\
\(\hat{x}_i\), \(\hat{x}\) & Generated time series. \\
\(L_i\), \(L\) & Series length. \\
\(\pi_\theta\) & Code-generation policy. \\
\(y\), \(y_j\) & Generated code and reference code. \\
\(\mathcal{Y}\) & Code space. \\
\(\mathrm{Execute}(\cdot)\) & Sandboxed code execution. \\
\(a_j\) & Structured temporal attribute. \\
\(a_j^{m}\) & Temporal factor, where \(m\) denotes length, trend, seasonality, event, or noise. \\
\(\mathcal{A}\) & Attribute space. \\
\(\mathcal{G}_{\mathrm{text}}\), \(\mathcal{G}_{\mathrm{code}}\) & Text/code generators. \\
\(\mathcal{D}_{\mathrm{syn}}\), \(N_s\) & Synthetic triplet dataset and its size. \\
\(R_{\mathrm{fmt}}\), \(R_{\mathrm{exec}}\), \(R_{\mathrm{ts}}\) & Format, execution, and TS rewards in the main formulation. \\
\(R_{\mathrm{len}}\), \(R_{\mathrm{err}}\), \(R_{\mathrm{corr}}\), \(R_{\mathrm{stat}}\) & Length, error, correlation, and statistic rewards. \\
\(\lambda_{\mathrm{fmt}},\lambda_{\mathrm{exec}},\lambda_{\mathrm{ts}}\) & Weights for format, execution, and TS rewards. \\
\(\lambda_{\mathrm{len}},\lambda_{\mathrm{err}},\lambda_{\mathrm{corr}},\lambda_{\mathrm{stat}}\) & Weights for TS reward components. \\
\(\mathbb{I}_{\mathrm{valid}}(y)\) & Indicator of parseable and executable generated code. \\
\(G\) & Candidates per prompt. \\
\(R_i^{(k)}\) & Reward of candidate \(k\). \\
\(\hat{A}_i^{(k)}\) & Group-normalized advantage. \\
\(\rho_{i,t}^{(k)}(\theta)\) & Token-level policy ratio. \\
\(\ell_i^{(k)}(\theta)\) & Clipped surrogate term for candidate \(k\). \\
\(T_k\) & Output-code length. \\
\(\varepsilon\) & Stability constant and clipping threshold. \\
\(\beta\) & KL coefficient. \\
\(D_{\mathrm{KL}}\) & Kullback-Leibler divergence. \\
\(\pi_{\mathrm{SFT}}\) & SFT reference policy. \\
\(\mathcal{L}_{\mathrm{SFT}}(\theta)\) & Supervised fine-tuning objective. \\
\(\mathcal{J}_{\mathrm{GRPO}}(\theta)\) & GRPO objective. \\
\(\theta_{\mathrm{SFT}}\), \(\theta_{\mathrm{final}}\) & SFT-initialized and final model parameters. \\
\bottomrule
\end{tabular}
\end{center}
\FloatBarrier

\section{Normalized Code Representation}
\label{app:normalized_code}

\begin{figure}[!t]
\centering
\includegraphics[width=\linewidth,trim=18pt 10pt 19pt 10pt,clip]{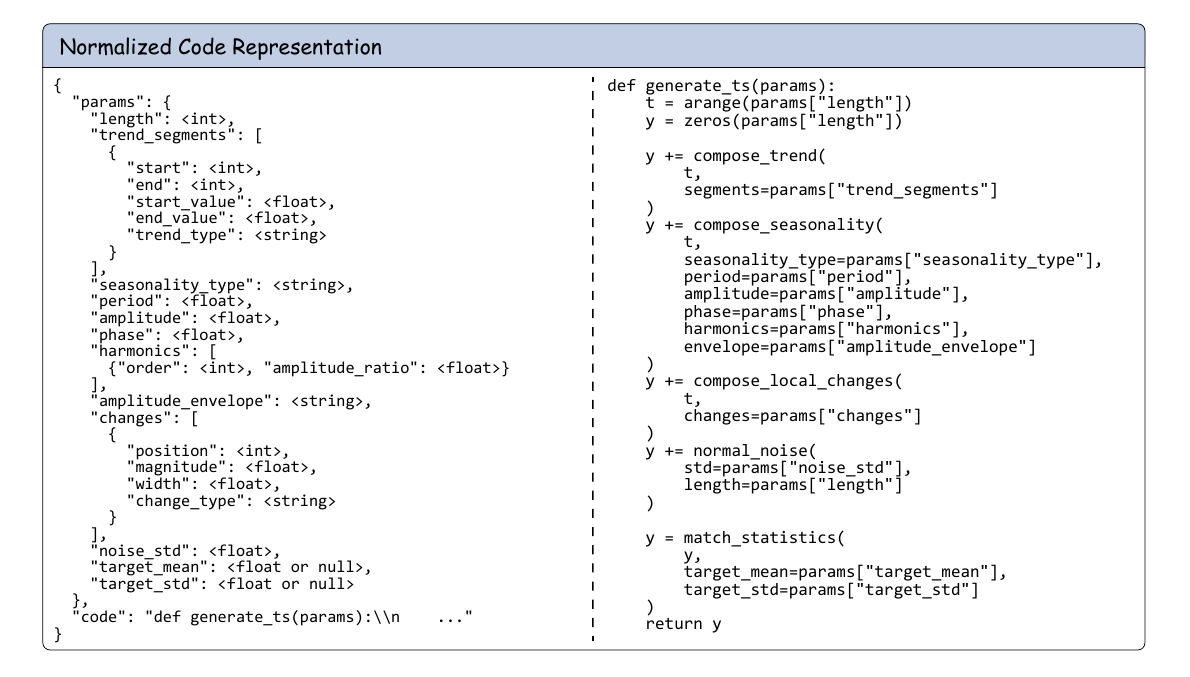}
\caption{Normalized code representation in CodeTS. The output separates
structured temporal parameters from the executable \texttt{generate\_ts}
function, enabling both format-level verification and execution-based time
series evaluation.}
\label{fig:norm_code}
\end{figure}

Figure~\ref{fig:norm_code} illustrates the normalized code representation used by CodeTS. Each output is a JSON object with two fields: \texttt{params} stores structured temporal specifications such as length, trend, seasonality, local changes, noise, and target statistics, while \texttt{code} stores an escaped Python function \texttt{generate\_ts(params)} that maps these parameters to a finite one-dimensional time series. This fixed interface keeps outputs parseable, executable, and rewardable, while still allowing diverse implementations of temporal patterns inside the generated code.

\FloatBarrier

\section{Detailed Reward Function Design}
\label{app:reward_details}

This section provides the concrete reward construction used in RLVR. The main
paper presents the reward at a high level as a combination of format validity,
executability, and time-series quality. In implementation, all reward terms are
scaled to \([0,1]\), and we expand Eq.~\eqref{eq:overall_reward} by absorbing
\(\lambda_{\mathrm{ts}}\) into the four time-series component weights:
\begin{equation}
\begin{aligned}
R(y,x,d) =
&\, \lambda_{\mathrm{fmt}} R_{\mathrm{fmt}}(y)
+ \lambda_{\mathrm{exec}} R_{\mathrm{exec}}(y) \\
&+ \mathbb{I}_{\mathrm{valid}}(y)
\left(
\lambda_{\mathrm{len}} R_{\mathrm{len}}(\hat{x},x)
+ \lambda_{\mathrm{err}} R_{\mathrm{err}}(\hat{x},x)\right.\\
&\left.\qquad\qquad
+ \lambda_{\mathrm{corr}} R_{\mathrm{corr}}(\hat{x},x)
+ \lambda_{\mathrm{stat}} R_{\mathrm{stat}}(\hat{x},x)
\right),
\end{aligned}
\label{eq:app_reward_full}
\end{equation}
where \(y\) is the generated JSON object, \(\hat{x}=\mathrm{Execute}(y)\) is
the executed time series, and \(\mathbb{I}_{\mathrm{valid}}(y)=1\) only when the
code passes parsing, sandbox execution, and returned-array validity checks.
The weights are set to \(\lambda_{\mathrm{fmt}}=0.05\),
\(\lambda_{\mathrm{exec}}=0.10\), \(\lambda_{\mathrm{len}}=0.10\),
\(\lambda_{\mathrm{err}}=0.30\), \(\lambda_{\mathrm{corr}}=0.30\), and
\(\lambda_{\mathrm{stat}}=0.15\).

\textbf{Format reward.}
\(R_{\mathrm{fmt}}\) gives graded feedback before execution. It checks whether
the model output follows the normalized \texttt{params}/\texttt{code}
interface:
\begin{equation}
R_{\mathrm{fmt}}(y)=
\begin{cases}
0.00, & \text{no JSON-like structure is detected},\\
0.20, & \text{JSON-like content is detected but parsing fails},\\
0.40, & \text{parsing succeeds but the result is not a dictionary},\\
0.60, & \text{the dictionary misses either \texttt{params} or \texttt{code}},\\
0.80, & \text{both fields exist but \texttt{generate\_ts(params)} is absent},\\
1.00, & \text{the normalized format is complete}.
\end{cases}
\label{eq:app_format_reward}
\end{equation}
This term reduces reward sparsity by distinguishing malformed text, incomplete
JSON, and complete normalized outputs.

\textbf{Execution reward.}
\(R_{\mathrm{exec}}\) evaluates whether the generated code can be compiled,
sandboxed, invoked, and converted into a valid numerical sequence:
\begin{equation}
R_{\mathrm{exec}}(y)=
\begin{cases}
0.00, & \text{compilation fails because of syntax errors},\\
0.25, & \text{syntax is valid but execution or function definition fails},\\
0.50, & \text{\texttt{generate\_ts(params)} is defined but invocation fails},\\
0.75, & \text{invocation succeeds but the returned value is invalid},\\
1.00, & \text{a finite one-dimensional numerical sequence is returned}.
\end{cases}
\label{eq:app_execution_reward}
\end{equation}
Invalid returned values include non-numerical objects, arrays with incompatible
shape, non-finite values, or empty sequences.

\textbf{Length reward.}
For a valid executed output, length consistency is scored by
\begin{equation}
R_{\mathrm{len}}(\hat{x},x)
=
\frac{\min(|\hat{x}|,|x|)}{\max(|\hat{x}|,|x|)}.
\label{eq:app_length_reward}
\end{equation}
This term provides partial credit for approximately correct output length while
penalizing sequences that are too short or too long.

\textbf{Pointwise error reward.}
Let \(T=\min(|\hat{x}|,|x|)\), and let \(\tilde{x}\) and
\(\tilde{\hat{x}}\) denote the first \(T\) points of \(x\) and \(\hat{x}\)
after per-sample min-max normalization. The error reward is
\begin{equation}
R_{\mathrm{err}}(\hat{x},x)
=
\max\left(1-\frac{1}{T}\sum_{t=1}^{T}
(\tilde{x}_t-\tilde{\hat{x}}_t)^2,\;0\right).
\label{eq:app_error_reward}
\end{equation}
The clipping prevents large numerical mismatches from producing negative
rewards.

\textbf{Correlation reward.}
The correlation term measures global shape agreement:
\begin{equation}
R_{\mathrm{corr}}(\hat{x},x)
=
\max\left(\rho(\tilde{\hat{x}},\tilde{x}),\;0\right),
\label{eq:app_corr_reward}
\end{equation}
where \(\rho(\cdot,\cdot)\) is Pearson correlation. If either normalized
sequence is numerically degenerate, this term is set to 0. Negative
correlations are also truncated to 0 because they indicate mismatched temporal
direction.

\textbf{Statistical reward.}
The statistical term is implemented as a standard-deviation ratio:
\begin{equation}
R_{\mathrm{stat}}(\hat{x},x)
=
\min\left(
\frac{\sigma(\hat{x}_{1:T})}{\sigma(x_{1:T})+\varepsilon_{\mathrm{stat}}},\;1
\right),
\label{eq:app_stat_reward}
\end{equation}
where \(\varepsilon_{\mathrm{stat}}\) is a small stability constant for the
standard-deviation ratio. This term directly penalizes low-variance collapsed
outputs, which may otherwise obtain nontrivial pointwise error or correlation
rewards after normalization. Excessive variance is mainly controlled by the
pointwise error term.

\section{Data Sources}
\label{app:Data_Sources}

\subsection{Text-Conditioned Benchmark Construction}
\label{app:benchmark_construction}

The eight evaluation benchmarks used in the main experiments are public time
series datasets that are originally released as numerical sequences rather than
as Text-TS pairs. To evaluate Text-to-TS generation, we therefore convert each
benchmark into a text-conditioned benchmark while keeping the original numerical
windows as the ground-truth targets. This conversion does not introduce
human-written captions or reference generation code; it only derives a natural
language description from each target window through a deterministic signal
analysis pipeline.

\textbf{Window construction.}
For each benchmark, we read the raw CSV file, automatically identify the time
column when present, and treat each remaining numerical column as a univariate
series. For each target length \(L\in\{96,192,336\}\), we segment every
univariate series into fixed-length windows using a stride of \(L/2\), i.e.,
48, 96, and 168 for the three generation lengths. Windows containing non-finite
values are discarded. Each retained window \(x\in\mathbb{R}^{L}\) becomes the
target sequence in one evaluation pair. The variable or column name is retained
as lightweight metadata and may appear in the generated description.

\textbf{Deterministic description extraction.}
For each target window, we extract temporal attributes directly from its
numerical values. The extractor first records summary statistics, including
length, mean, standard deviation, minimum, and maximum. It then estimates trend
with linear regression, optionally using a two-segment trend when the segmented
fit significantly improves over a single global line. After removing the trend,
it detects dominant seasonality from the residual using FFT-based spectral
analysis, with an autocorrelation fallback for short or ambiguous windows. The
seasonal component is summarized by period, amplitude, phase, number of visible
cycles, and amplitude-envelope type. Local events are then detected as peaks and
troughs on the residual after removing trend and seasonality; each event records
its direction, start position, turning point, end position, magnitude, and
whether it is spike-like or smooth. The remaining residual variance is reported
as the noise level.

\textbf{Text-TS pair formation.}
The extracted attributes are rendered into a fixed natural-language template.
For example, a description may state the sequence identifier, the length and
summary statistics, the trend direction and endpoints, the dominant seasonal
period and amplitude, a list of local events, and the residual noise level. The
final evaluation example is therefore
\[
(d,x),
\]
where \(d\) is the automatically rendered description and \(x\) is the original
benchmark window. During evaluation, a method receives only \(d\) and must
generate a time series \(\hat{x}\); all reported metrics compare \(\hat{x}\)
against the unchanged original numerical window \(x\). For supervised baselines,
the same conversion procedure is applied to the corresponding training split of
each target benchmark. For CodeTS and zero-shot LLM baselines, only the
converted test descriptions are used at inference time.

This construction makes the evaluation compatible with Text-to-TS generation
while preserving the original public benchmark values. It should be interpreted
as a text-conditioned synthesis benchmark: the text describes the target
temporal behavior derived from the target window, and the task is to realize
that described behavior as a numerical sequence, rather than to forecast future
values from historical context.

\subsection{Real-World Training Data Sources}
\label{app:real_data_sources}
Table~\ref{tab:real_data_sources} lists the real-world data sources used to construct the real Text-TS pairs for RLVR. The entries are filtered according to the final RLVR training file and linked to the corresponding public data-source pages. These sources are used only for training-data construction and are kept separate from the evaluation benchmarks reported in the main experiments.

\begin{table}[!t]
\centering
\caption{Real-world data sources used for RLVR training.}
\label{tab:real_data_sources}
\footnotesize
\setlength{\tabcolsep}{4pt}
\setlength{\Urlmuskip}{0mu plus 1mu}
\renewcommand{\arraystretch}{1.10}
\begin{tabular}{p{0.34\linewidth}p{0.58\linewidth}}
\toprule
Dataset & URL \\
\midrule
Alibaba Cluster Trace & \url{https://github.com/alibaba/clusterdata} \\
Binance BTC/USDT & \url{https://data.binance.vision/} \\
Currency Hourly/Volatility & \url{https://finance.yahoo.com/currencies/} \\
ESC-50 & \url{https://github.com/karolpiczak/ESC-50} \\
Freesound Dataset 50k (FSD50K) & \url{https://zenodo.org/records/4060432} \\
GDELT Event Data & \url{https://www.gdeltproject.org/data.html} \\
Google COVID-19 Open Data & \url{https://github.com/GoogleCloudPlatform/covid-19-open-data} \\
IMS Bearing Data (NASA) & \url{https://www.nasa.gov/intelligent-systems-division/discovery-and-systems-health/pcoe/pcoe-data-set-repository/} \\
MIT-BIH Arrhythmia & \url{https://physionet.org/content/mitdb/1.0.0/} \\
M5 Forecasting (Walmart) & \url{https://www.kaggle.com/c/m5-forecasting-accuracy} \\
Natural Gas Futures & \url{https://www.eia.gov/dnav/ng/ng_pri_fut_s1_d.htm} \\
NREL Wind Toolkit & \url{https://developer.nrel.gov/docs/wind/wind-toolkit/wtk-download/} \\
NYC Yellow Taxi Trip Data & \url{https://www.nyc.gov/site/tlc/about/tlc-trip-record-data.page} \\
OpenMeteo Solar Data & \url{https://open-meteo.com/en/docs/historical-weather-api} \\
Paderborn University Bearing Dataset & \url{https://mb.uni-paderborn.de/en/kat/research/bearing-datacenter/data-sets-and-download} \\
PTB-XL & \url{https://physionet.org/content/ptb-xl/1.0.3/} \\
UK-DALE & \url{https://jack-kelly.com/data/} \\
USGS Daily Streamflow & \url{https://api.waterdata.usgs.gov/ogcapi/v0/collections/daily} \\
\bottomrule
\end{tabular}
\end{table}
\FloatBarrier

\section{Evaluation Metrics and Success Rate}
\label{app:metrics}

Let \(\mathcal{V}\) denote the valid sample index set after removing
near-constant target windows and clearly invalid predictions. For each
\(i\in\mathcal{V}\), we align \((x_i,\hat{x}_i)\) by truncation to
\(T_i=\min(|x_i|,|\hat{x}_i|)\). Let
\(m_i=\min_{1\le t\le T_i}x_{i,t}\) and
\(M_i=\max_{1\le t\le T_i}x_{i,t}\). The normalized sequences are
\begin{equation}
\tilde{x}_{i,t}
=
\frac{x_{i,t}-m_i}{M_i-m_i},
\qquad
\tilde{\hat{x}}_{i,t}
=
\frac{\hat{x}_{i,t}-m_i}{M_i-m_i},
\qquad
t=1,\ldots,T_i .
\label{eq:app_eval_norm}
\end{equation}
Samples with numerically degenerate \(M_i-m_i\) are excluded from
\(\mathcal{V}\), and the same filtering is applied to all methods.

The reported MSE is
\begin{equation}
\mathrm{MSE}
=
\frac{1}{|\mathcal{V}|}
\sum_{i\in\mathcal{V}}
\frac{1}{T_i}
\sum_{t=1}^{T_i}
\left(\tilde{x}_{i,t}-\tilde{\hat{x}}_{i,t}\right)^2 .
\label{eq:app_eval_mse}
\end{equation}

The reported DTW is the averaged normalized warping cost:
\begin{equation}
\mathrm{DTW}
=
\frac{1}{|\mathcal{V}|}
\sum_{i\in\mathcal{V}}
\frac{1}{T_i}
\min_{\pi\in\Pi_i}
\sum_{(p,q)\in\pi}
\left|\tilde{x}_{i,p}-\tilde{\hat{x}}_{i,q}\right| ,
\label{eq:app_eval_dtw}
\end{equation}
where \(\Pi_i\) is the set of monotone warping paths between the two aligned
sequences.

Pearson correlation is averaged as
\begin{equation}
\mathrm{PC}
=
\frac{1}{|\mathcal{V}|}
\sum_{i\in\mathcal{V}}
\frac{
\sum_{t=1}^{T_i}
(\tilde{x}_{i,t}-\bar{x}_i)
(\tilde{\hat{x}}_{i,t}-\bar{\hat{x}}_i)
}{
\sqrt{\sum_{t=1}^{T_i}(\tilde{x}_{i,t}-\bar{x}_i)^2}
\sqrt{\sum_{t=1}^{T_i}(\tilde{\hat{x}}_{i,t}-\bar{\hat{x}}_i)^2}
},
\label{eq:app_eval_pc}
\end{equation}
where \(\bar{x}_i\) and \(\bar{\hat{x}}_i\) are means of the normalized
sequences.

For LLM-based zero-shot baselines, we report success rate as pass@\(k\)
\citep{chen2021evaluating}. For task \(i\), suppose \(n_i\) candidates are
sampled and \(c_i\) of them are successful. The unbiased estimator is
\begin{equation}
\widehat{\mathrm{pass@}k}
=
\frac{1}{N}
\sum_{i=1}^{N}
\left(
1-
\frac{\binom{n_i-c_i}{k}}{\binom{n_i}{k}}
\right),
\qquad
n_i\ge k .
\label{eq:app_pass_at_k}
\end{equation}
Here \(c_i=\sum_{j=1}^{n_i}s_{i,j}\), where \(s_{i,j}=1\) if candidate \(j\)
passes all required checks and \(0\) otherwise. For code-generating methods,
these checks require JSON parsing, the \texttt{params}/\texttt{code} fields,
the \texttt{generate\_ts(params)} function, sandboxed execution, and a finite
one-dimensional returned sequence with compatible length. For textualized
baselines, the parsing and shape checks are adapted to their output format. The
reported SR is the pass@1 special case:
\begin{equation}
\mathrm{SR}(\%)
=
100\times \widehat{\mathrm{pass@}1}.
\label{eq:app_success_rate}
\end{equation}

\section{Additional Comparison with Recent LLMs under Text-to-Code-to-TS}
\label{app:code_llm_baselines}

We further evaluate several mainstream and recent strong LLMs under the same Text-to-Code-to-TS prompting setup, including IQuest-Coder-V1-7B-Instruct~\citep{yang2026iquest}, Devstral-Instruct-24B~\citep{rastogi2025devstral}, Seed-Coder-8B-Instruct~\citep{seed2025seed}, and Qwen3.5-9B~\citep{team2026qwen3}. These models cover recent code-oriented LLMs, a larger coding-capable instruction model, and a strong general-purpose LLM with coding capability. This comparison examines whether the gains of CodeTS come from the proposed training framework rather than from code generation ability alone.

As shown in Table~\ref{tab:comparable_llm_scale}, CodeTS achieves the strongest overall performance across datasets and generation lengths. Although some prompted LLMs obtain competitive results on a few individual metrics, their generated series generally show weaker temporal fidelity and less stable execution success. In contrast, CodeTS maintains consistently low MSE and DTW, substantially higher Pearson correlation, and near-perfect pass@1 success rate. These results indicate that the advantage of CodeTS does not come merely from prompting a strong LLM to write code, but from learning the normalized Text-to-Code-to-TS interface and optimizing generation quality through execution-based feedback.

\begin{table*}[htbp]
\Huge
\centering
\setlength{\abovecaptionskip}{2pt}
\setlength{\belowcaptionskip}{1pt}
\caption{Results of mainstream and recent strong coding-capable LLMs under the Text-to-Code-to-TS setup. SR denotes the pass@1 success rate.}
\label{tab:comparable_llm_scale}
\setlength{\tabcolsep}{3.0pt}
\renewcommand{\arraystretch}{0.92}
\begin{adjustbox}{max width=\textwidth,max totalheight=0.42\textheight,center}
\begin{tabular}{ll|cccc|cccc|cccc|cccc|cccc}
\toprule
 & & \multicolumn{4}{c|}{CodeTS (ours) (7B)} & \multicolumn{4}{c|}{IQuest-Coder-V1-7B-Instruct (7B)} & \multicolumn{4}{c|}{Seed-Coder-8B-Instruct (8B)} & \multicolumn{4}{c|}{Qwen3.5-9B (9B)} & \multicolumn{4}{c}{Devstral-Instruct-24B (24B)} \\
\cmidrule(lr){3-6}\cmidrule(lr){7-10}\cmidrule(lr){11-14}\cmidrule(lr){15-18}\cmidrule(lr){19-22}
Datasets & Length & MSE$\downarrow$ & DTW$\downarrow$ & Pearson$\uparrow$ & SR(\%)$\uparrow$ & MSE$\downarrow$ & DTW$\downarrow$ & Pearson$\uparrow$ & SR(\%)$\uparrow$ & MSE$\downarrow$ & DTW$\downarrow$ & Pearson$\uparrow$ & SR(\%)$\uparrow$ & MSE$\downarrow$ & DTW$\downarrow$ & Pearson$\uparrow$ & SR(\%)$\uparrow$ & MSE$\downarrow$ & DTW$\downarrow$ & Pearson$\uparrow$ & SR(\%)$\uparrow$ \\
\midrule
\multirow{3}{*}{ETTh1} & 96 & \textbf{0.0202} & \textbf{0.0799} & \textbf{0.8136} & 100.0 & 0.1902 & 0.1815 & 0.1463 & 96.3 & 0.0975 & 0.1338 & 0.1632 & 99.2 & \underline{0.0910} & \underline{0.1316} & \underline{0.2201} & 94.1 & 0.0915 & 0.1324 & 0.2172 & 98.4 \\
 & 192 & \textbf{0.0219} & \textbf{0.0808} & \textbf{0.7714} & 100.0 & 0.1516 & 0.1587 & 0.1161 & 92.5 & 0.0878 & 0.1261 & 0.1360 & 95.6 & \underline{0.0845} & 0.1256 & \underline{0.1742} & 88.5 & 0.0867 & \underline{0.1249} & 0.1550 & 97.6 \\
 & 336 & \textbf{0.0232} & \textbf{0.0823} & \textbf{0.7242} & 100.0 & 0.1116 & 0.1500 & 0.0969 & 89.1 & 0.0798 & 0.1250 & 0.1195 & 95.9 & \underline{0.0787} & 0.1251 & \underline{0.1424} & 83.7 & 0.0789 & \underline{0.1242} & 0.1288 & 97.3 \\
\midrule
\multirow{3}{*}{ETTh2} & 96 & \textbf{0.0305} & \textbf{0.1041} & \textbf{0.7302} & 100.0 & 0.1870 & 0.1860 & 0.1508 & 93.3 & 0.0926 & 0.1436 & 0.1689 & 98.0 & 0.0881 & 0.1436 & \underline{0.2123} & 93.7 & \underline{0.0876} & \underline{0.1419} & 0.2111 & 97.8 \\
 & 192 & \textbf{0.0295} & \textbf{0.0999} & \textbf{0.6835} & 100.0 & 0.1025 & 0.1512 & 0.1556 & 91.3 & 0.0793 & \underline{0.1345} & 0.1572 & 96.4 & 0.0828 & 0.1404 & \underline{0.1804} & 84.9 & \underline{0.0777} & 0.1351 & 0.1767 & 96.8 \\
 & 336 & \textbf{0.0292} & \textbf{0.0993} & \textbf{0.6456} & 100.0 & 0.0983 & 0.1503 & 0.1474 & 89.8 & 0.0711 & \underline{0.1322} & 0.1510 & 100.0 & 0.0752 & 0.1373 & 0.1626 & 85.0 & \underline{0.0689} & 0.1337 & \underline{0.1771} & 95.9 \\
\midrule
\multirow{3}{*}{ETTm1} & 96 & \textbf{0.0161} & \textbf{0.0669} & \textbf{0.8713} & 100.0 & 0.2059 & 0.1964 & 0.4431 & 93.8 & 0.0649 & 0.1258 & 0.4973 & 97.8 & 0.0595 & 0.1226 & \underline{0.5471} & 90.2 & \underline{0.0588} & \underline{0.1219} & 0.5459 & 98.2 \\
 & 192 & \textbf{0.0181} & \textbf{0.0685} & \textbf{0.8338} & 100.0 & 0.1942 & 0.1811 & 0.2702 & 94.4 & 0.0836 & 0.1256 & 0.2886 & 95.7 & \underline{0.0798} & 0.1241 & \underline{0.3293} & 87.5 & 0.0809 & \underline{0.1233} & 0.3223 & 96.2 \\
 & 336 & \textbf{0.0276} & \textbf{0.0704} & \textbf{0.7341} & 100.0 & 0.1299 & 0.1716 & 0.1217 & 92.4 & 0.0901 & \underline{0.1414} & 0.1365 & 95.4 & \underline{0.0841} & 0.1427 & \underline{0.1916} & 81.6 & 0.0865 & 0.1436 & 0.1759 & 97.1 \\
\midrule
\multirow{3}{*}{ETTm2} & 96 & \textbf{0.0279} & \textbf{0.0982} & \textbf{0.7944} & 100.0 & 0.1760 & 0.1867 & 0.3395 & 92.8 & 0.0860 & 0.1499 & 0.3535 & 97.4 & \underline{0.0789} & 0.1460 & \underline{0.4042} & 89.3 & 0.0796 & \underline{0.1456} & 0.4023 & 98.0 \\
 & 192 & \textbf{0.0295} & \textbf{0.1016} & \textbf{0.7532} & 100.0 & 0.1373 & 0.1731 & 0.2345 & 94.8 & 0.0873 & 0.1437 & 0.2433 & 96.4 & 0.0822 & 0.1425 & \underline{0.2887} & 87.6 & \underline{0.0817} & \underline{0.1414} & 0.2873 & 96.8 \\
 & 336 & \textbf{0.0332} & \textbf{0.0968} & \textbf{0.6689} & 100.0 & 0.1206 & 0.1728 & 0.1403 & 92.6 & 0.0857 & \underline{0.1449} & 0.1422 & 95.5 & 0.0859 & 0.1492 & 0.1701 & 80.7 & \underline{0.0826} & 0.1460 & \underline{0.1708} & 98.5 \\
\midrule
\multirow{3}{*}{Weather} & 96 & \textbf{0.0159} & \textbf{0.0622} & \textbf{0.8599} & 100.0 & 0.1629 & 0.1626 & 0.4714 & 93.5 & 0.0685 & 0.1320 & 0.4869 & 97.1 & 0.0621 & 0.1268 & \underline{0.5296} & 83.4 & \underline{0.0617} & \underline{0.1259} & 0.5282 & 97.6 \\
 & 192 & \textbf{0.0159} & \textbf{0.0592} & \textbf{0.8427} & 100.0 & 0.1341 & 0.1527 & 0.4296 & 94.3 & 0.0718 & 0.1304 & 0.4279 & 96.7 & 0.0663 & 0.1262 & 0.4662 & 81.0 & \underline{0.0646} & \underline{0.1243} & \underline{0.4730} & 97.8 \\
 & 336 & \textbf{0.0213} & \textbf{0.0648} & \textbf{0.7701} & 100.0 & 0.1277 & 0.1623 & 0.2729 & 95.1 & 0.0778 & \underline{0.1328} & 0.2914 & 96.9 & \underline{0.0730} & 0.1333 & \underline{0.3234} & 81.1 & 0.0739 & 0.1328 & 0.3158 & 97.5 \\
\midrule
\multirow{3}{*}{Electricity} & 96 & \textbf{0.0301} & \textbf{0.0923} & \textbf{0.8112} & 100.0 & 0.1809 & 0.1599 & 0.1139 & 96.4 & 0.1610 & \underline{0.1463} & 0.1133 & 97.7 & 0.1593 & 0.1468 & 0.1169 & 90.1 & \underline{0.1589} & 0.1476 & \underline{0.1220} & 97.4 \\
 & 192 & \textbf{0.0284} & \textbf{0.0845} & \textbf{0.7982} & 99.8 & 0.1686 & 0.1463 & \underline{0.0498} & 93.7 & 0.1553 & 0.1311 & 0.0470 & 97.4 & 0.1558 & 0.1312 & 0.0459 & 89.9 & \underline{0.1549} & \underline{0.1310} & 0.0483 & 97.3 \\
 & 336 & \textbf{0.0280} & \textbf{0.0824} & \textbf{0.7811} & 99.9 & 0.1595 & 0.1391 & \underline{0.0257} & 88.8 & 0.1448 & \underline{0.1260} & 0.0211 & 97.7 & 0.1453 & 0.1260 & 0.0212 & 86.9 & \underline{0.1437} & 0.1261 & 0.0213 & 94.7 \\
\midrule
\multirow{3}{*}{Exchange} & 96 & \textbf{0.0795} & \textbf{0.1643} & \underline{0.4815} & 100.0 & 0.4000 & 0.2341 & \textbf{0.4954} & 97.3 & 0.1229 & 0.2045 & 0.1251 & 96.7 & \underline{0.0932} & 0.1828 & 0.3344 & 95.1 & 0.0983 & \underline{0.1804} & 0.3100 & 96.7 \\
 & 192 & \underline{0.0837} & \textbf{0.1481} & \textbf{0.5180} & 100.0 & 0.3163 & 0.2181 & \underline{0.4745} & 92.3 & 0.1018 & 0.1723 & 0.1820 & 91.2 & \textbf{0.0757} & \underline{0.1505} & 0.3913 & 91.2 & 0.0866 & 0.1505 & 0.2901 & 97.8 \\
 & 336 & 0.0870 & 0.1449 & \textbf{0.5159} & 100.0 & 0.2291 & 0.2051 & \underline{0.4343} & 98.2 & 0.0978 & 0.1604 & 0.1211 & 98.2 & \textbf{0.0745} & \underline{0.1402} & 0.3574 & 87.5 & \underline{0.0768} & \textbf{0.1330} & 0.2956 & 100.0 \\
\midrule
\multirow{3}{*}{Web} & 96 & \textbf{0.0326} & \underline{0.1162} & \textbf{0.3672} & 100.0 & 0.0954 & 0.1489 & 0.0921 & 94.9 & 0.0501 & 0.1168 & 0.0689 & 97.7 & 0.0491 & 0.1162 & \underline{0.1001} & 74.7 & \underline{0.0482} & \textbf{0.1153} & 0.0975 & 97.8 \\
 & 192 & \textbf{0.0246} & 0.1002 & \textbf{0.3024} & 100.0 & 0.0688 & 0.1367 & 0.0682 & 95.9 & 0.0357 & 0.0998 & 0.0451 & 97.3 & 0.0349 & \underline{0.0991} & \underline{0.0756} & 69.9 & \underline{0.0343} & \textbf{0.0981} & 0.0690 & 97.4 \\
 & 336 & \textbf{0.0175} & \textbf{0.0823} & \textbf{0.2664} & 99.9 & 0.0570 & 0.1310 & \underline{0.0545} & 96.2 & 0.0245 & 0.0845 & 0.0339 & 97.9 & 0.0239 & 0.0837 & 0.0518 & 68.4 & \underline{0.0235} & \underline{0.0829} & 0.0436 & 96.7 \\
\bottomrule
\end{tabular}
\end{adjustbox}
\vspace{-0.6em}
\end{table*}

\section{Training Configuration}
\label{app:training_config}

This section supplements the details of the implementation of the two-stage training procedure, focusing on model initialization, sample organization, parallel training strategy, and optimization hyperparameters.

\subsection{SFT Warmup Configuration}

In the SFT warm-up stage, we initialize the policy model from Qwen2.5-Coder-7B-Instruct and conduct full-parameter supervised fine-tuning using DeepSpeed on four NVIDIA A100-SXM4-80GB GPUs. 
Each training instance is formatted as a dialogue-style sequence, where the input consists of the task instruction and the corresponding textual description, and the output is the target JSON object. 
We adopt the chat template associated with the base model and pack multiple short instances into a single training sequence whenever feasible to improve computational efficiency.

This stage is conducted with 500 training examples for one epoch, using a maximum sequence length of 3072. 
The global batch size is set to 64, with a per-device micro-batch size of 8. 
We use a learning rate of \(5\times10^{-6}\), a warm-up ratio of 0.05, and a cosine learning-rate schedule with a nonzero minimum learning rate. 
Training is performed in BF16 precision. 
To reduce memory consumption, we employ ZeRO-2 optimization together with gradient checkpointing. 
During training, only the most recent checkpoint is retained, and the final model is exported in the HuggingFace format.

This configuration reflects the intended function of the warm-up stage in our framework. Rather than aiming to overfit the synthetic reference code, this stage is designed to provide the model with an initial alignment to the expected output format, parameter organization, and executable code scaffold, while incurring only a limited computational cost. The deliberately short training schedule and small training set further preserve sufficient exploration capacity for the subsequent RLVR stage.

\subsection{GRPO Configuration}

In the RLVR stage, we continue training from the SFT warm-up model using a training pipeline built on OpenRLHF, Ray, and vLLM. 
This stage is conducted on two NVIDIA A100-SXM4-80GB GPUs. 
In contrast to the SFT warm-up stage, the training data consist of real text--time-series pairs. 
The input side contains a textual description, while the target side provides the corresponding time series and its length information, which are used to compute verifiable rewards from executed code outputs. 
We use the chat template associated with the base model, and set the maximum input length and maximum generation length to 1024 and 3072, respectively.

The RLVR stage uses 6,300 training examples. 
For each input, we sample four candidate programs, corresponding to \(G=4\) in the main paper. 
Both the rollout batch size and the training batch size are set to 16, while the micro rollout batch size and micro training batch size are both set to 2. 
The model is trained for one epoch and one episode. 
For optimization, the actor learning rate is set to \(5\times10^{-7}\), the initial KL coefficient is \(10^{-3}\), and the clipping range is \([\varepsilon_{\text{low}}, \varepsilon_{\text{high}}]=[0.2, 0.28]\). 
Advantage estimation is performed with group normalization. 
Training is conducted in BF16 precision. 
To reduce memory consumption, we employ ZeRO-3 optimization, Adam offloading, and gradient checkpointing. 
Online sampling is performed with a single vLLM engine using tensor parallel size 2 and a GPU memory utilization ratio of 0.45. 
The actor model, reference policy, and rollout engine are deployed in colocated mode and synchronized through NCCL.

A central consideration in this stage is the resource balance among online sampling, reward evaluation, and gradient-based policy updates. The relatively conservative rollout batch size, number of sampled candidates, and micro-batch configuration are chosen to limit the sampling pressure at each update step, avoid excessive reuse of the same real-data batch, and reserve sufficient memory for generating, executing, and scoring long-form code outputs.

\section{Web Benchmark Analysis}
\label{app:web_analysis}

The Web benchmark has a more event-concentrated structure than the other benchmarks. 
In the target windows used for evaluation, near-flat adjacent steps account for 13.5\%, 18.9\%, and 25.5\% of the normalized transitions at lengths 96, 192, and 336, respectively. 
At the same time, the largest 5\% of adjacent changes explain 30.9\%, 34.9\%, and 37.4\% of the total variation. 
This combination makes Web a stress test for preserving sparse shape changes rather than only matching the dominant low-variation regions.

\begin{figure}[t]
\centering
\includegraphics[width=\textwidth]{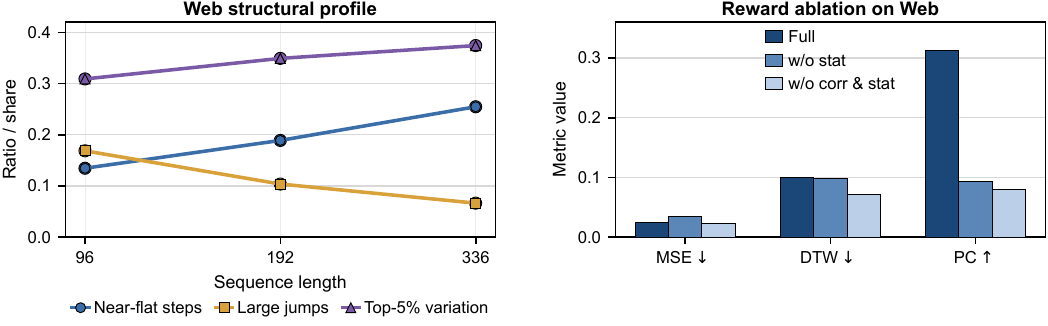}
\caption{Web benchmark analysis. Left: target-window statistics for Web, where near-flat steps are normalized adjacent differences no larger than 0.01, large jumps are differences at least 0.20, and top-5\% variation is the share of total variation contributed by the largest 5\% adjacent changes. Right: reward ablation on Web averaged over lengths 96, 192, and 336. Removing correlation and statistic rewards can reduce MSE and DTW while sharply reducing Pearson correlation, showing that low pointwise or warped error does not necessarily imply event-shape preservation on Web.}
\label{fig:web_benchmark_analysis}
\end{figure}

Figure~\ref{fig:web_benchmark_analysis} summarizes this behavior. 
The final CodeTS model achieves Pearson correlations of 0.367, 0.302, and 0.266 on Web for lengths 96, 192, and 336, respectively. 
However, the reward ablation shows that MSE and DTW alone are not sufficient to characterize this benchmark. 
After removing both the correlation and statistic rewards, Web MSE decreases from 0.0249 to 0.0230 and DTW decreases from 0.0996 to 0.0721, but Pearson drops from 0.3120 to 0.0803. 
Thus, a model can match many local values or warped segments while failing to preserve the sparse event-level shape. 
We therefore interpret Pearson correlation as the most diagnostic metric for Web, with MSE and DTW serving as complementary error measures.

\section{Detailed Ablation Studies}
\label{app:ablation_details}

\textbf{Detailed Training-Stage Ablation}
Table~\ref{tab:stage_ablation_detail} provides the detailed training stage ablation results for each benchmark dataset and generation length. Compared with the SFT-warmup model, the final SFT+RLVR model achieves substantially better generation quality in most settings, with lower MSE and DTW, higher Pearson correlation, and higher pass@1 success rate. These results further support the conclusion in the main paper that SFT provides a useful initialization for executable code generation, while RLVR is the key stage that improves the executed time series through verifiable feedback from real Text-TS pairs.

\begin{table*}[t]
\centering
\setlength{\abovecaptionskip}{2pt}
\setlength{\belowcaptionskip}{1pt}
\caption{Training-stage ablation; SR is pass@1 success rate.}
\label{tab:stage_ablation_detail}
\scriptsize
\setlength{\tabcolsep}{3.0pt}
\renewcommand{\arraystretch}{0.80}
\begin{tabular*}{\textwidth}{@{\extracolsep{\fill}}ll|cccc|cccc@{}}
\toprule
 & & \multicolumn{4}{c|}{SFT} & \multicolumn{4}{c}{SFT+RLVR} \\
\cmidrule(lr){3-10}
Datasets & Length & MSE$\downarrow$ & DTW$\downarrow$ & Pearson$\uparrow$ & SR(\%)$\uparrow$ & MSE$\downarrow$ & DTW$\downarrow$ & Pearson$\uparrow$ & SR(\%)$\uparrow$ \\
\midrule
\multirow{3}{*}{ETTh1} & 96 & \underline{0.0938} & \underline{0.1358} & \underline{0.1972} & 95.1 & \textbf{0.0202} & \textbf{0.0799} & \textbf{0.8136} & 100.0 \\
 & 192 & \underline{0.0909} & \underline{0.1342} & \underline{0.1155} & 87.7 & \textbf{0.0219} & \textbf{0.0808} & \textbf{0.7714} & 100.0 \\
 & 336 & \underline{0.0824} & \underline{0.1334} & \underline{0.1069} & 84.4 & \textbf{0.0232} & \textbf{0.0823} & \textbf{0.7242} & 100.0 \\
\midrule
\multirow{3}{*}{ETTh2} & 96 & \underline{0.0894} & \underline{0.1455} & \underline{0.1979} & 92.2 & \textbf{0.0305} & \textbf{0.1041} & \textbf{0.7302} & 100.0 \\
 & 192 & \underline{0.0776} & \underline{0.1361} & \underline{0.1707} & 86.1 & \textbf{0.0295} & \textbf{0.0999} & \textbf{0.6835} & 100.0 \\
 & 336 & \underline{0.0701} & \underline{0.1349} & \underline{0.1604} & 79.6 & \textbf{0.0292} & \textbf{0.0993} & \textbf{0.6456} & 100.0 \\
\midrule
\multirow{3}{*}{ETTm1} & 96 & \underline{0.0635} & \underline{0.1268} & \underline{0.5096} & 94.0 & \textbf{0.0161} & \textbf{0.0669} & \textbf{0.8713} & 100.0 \\
 & 192 & \underline{0.0820} & \underline{0.1299} & \underline{0.2973} & 88.3 & \textbf{0.0181} & \textbf{0.0685} & \textbf{0.8338} & 100.0 \\
 & 336 & \underline{0.0873} & \underline{0.1461} & \underline{0.1424} & 84.9 & \textbf{0.0276} & \textbf{0.0704} & \textbf{0.7341} & 100.0 \\
\midrule
\multirow{3}{*}{ETTm2} & 96 & \underline{0.0831} & \underline{0.1489} & \underline{0.3796} & 90.0 & \textbf{0.0279} & \textbf{0.0982} & \textbf{0.7944} & 100.0 \\
 & 192 & \underline{0.0849} & \underline{0.1450} & \underline{0.2757} & 90.0 & \textbf{0.0295} & \textbf{0.1016} & \textbf{0.7532} & 100.0 \\
 & 336 & \underline{0.0844} & \underline{0.1501} & \underline{0.1662} & 86.4 & \textbf{0.0332} & \textbf{0.0968} & \textbf{0.6689} & 100.0 \\
\midrule
\multirow{3}{*}{Weather} & 96 & \underline{0.0630} & \underline{0.1256} & \underline{0.5209} & 90.2 & \textbf{0.0159} & \textbf{0.0622} & \textbf{0.8599} & 100.0 \\
 & 192 & \underline{0.0690} & \underline{0.1280} & \underline{0.4469} & 89.6 & \textbf{0.0159} & \textbf{0.0592} & \textbf{0.8427} & 100.0 \\
 & 336 & \underline{0.0745} & \underline{0.1355} & \underline{0.3119} & 85.6 & \textbf{0.0213} & \textbf{0.0648} & \textbf{0.7701} & 100.0 \\
\midrule
\multirow{3}{*}{Electricity} & 96 & \underline{0.1590} & \underline{0.1534} & \underline{0.1193} & 90.9 & \textbf{0.0301} & \textbf{0.0923} & \textbf{0.8112} & 100.0 \\
 & 192 & \underline{0.1576} & \underline{0.1473} & \underline{0.0373} & 87.6 & \textbf{0.0284} & \textbf{0.0845} & \textbf{0.7982} & 99.8 \\
 & 336 & \underline{0.1466} & \underline{0.1430} & \underline{0.0139} & 85.7 & \textbf{0.0280} & \textbf{0.0824} & \textbf{0.7811} & 99.9 \\
\midrule
\multirow{3}{*}{Exchange} & 96 & \underline{0.1108} & \underline{0.1807} & \underline{0.2372} & 79.7 & \textbf{0.0795} & \textbf{0.1643} & \textbf{0.4815} & 100.0 \\
 & 192 & \underline{0.0941} & \underline{0.1514} & \underline{0.2716} & 70.3 & \textbf{0.0837} & \textbf{0.1481} & \textbf{0.5180} & 100.0 \\
 & 336 & \textbf{0.0819} & \textbf{0.1407} & \underline{0.3138} & 80.4 & \underline{0.0870} & \underline{0.1449} & \textbf{0.5159} & 100.0 \\
\midrule
\multirow{3}{*}{Web} & 96 & \underline{0.0486} & \textbf{0.1148} & \underline{0.0945} & 75.1 & \textbf{0.0326} & \underline{0.1162} & \textbf{0.3672} & 100.0 \\
 & 192 & \underline{0.0344} & \textbf{0.0970} & \underline{0.0675} & 68.8 & \textbf{0.0246} & \underline{0.1002} & \textbf{0.3024} & 100.0 \\
 & 336 & \underline{0.0230} & \textbf{0.0809} & \underline{0.0429} & 67.9 & \textbf{0.0175} & \underline{0.0823} & \textbf{0.2664} & 99.9 \\
\bottomrule
\end{tabular*}
\vspace{-0.6em}
\end{table*}

% \section{Complete Results}
% \label{app:complete_results}

% This section supplements the complete results that are not fully expanded in the main paper, including zero-shot inference comparisons among LLMs of comparable parameter scales under the Text-to-Code-to-TS framework, training stage ablations, reward design ablations, normalized code ablations, and template dependence analysis.

\textbf{Detailed Normalized Code Ablation}
% \label{app:normalized_code_ablation_details}
Table~\ref{tab:normalized_code_ablation_detail} reports the detailed normalized code ablation results for each benchmark dataset and generation length. Compared with the variant without normalized code, the full model generally achieves better generation quality, with lower MSE and DTW and higher Pearson correlation on most datasets. This confirms that the benefit of CodeTS comes not only from using executable code, but also from organizing the output into explicit temporal parameters and executable logic. The normalized representation provides a more structured interface for parsing, execution checking, and reward computation.

\begin{table*}[t]
\centering
\setlength{\abovecaptionskip}{2pt}
\setlength{\belowcaptionskip}{1pt}
\caption{Normalized code ablation; SR is pass@1 success rate.}
\label{tab:normalized_code_ablation_detail}
\scriptsize
\setlength{\tabcolsep}{3.0pt}
\renewcommand{\arraystretch}{0.80}
\begin{tabular*}{\textwidth}{@{\extracolsep{\fill}}ll|cccc|cccc@{}}
\toprule
 & & \multicolumn{4}{c|}{Full} & \multicolumn{4}{c}{w/o Normalized Code} \\
\cmidrule(lr){3-6}\cmidrule(lr){7-10}
Datasets & Length & MSE$\downarrow$ & DTW$\downarrow$ & Pearson$\uparrow$ & SR(\%)$\uparrow$ & MSE$\downarrow$ & DTW$\downarrow$ & Pearson$\uparrow$ & SR(\%)$\uparrow$ \\
\midrule
\multirow{3}{*}{ETTh1} & 96 & \textbf{0.0202} & \textbf{0.0799} & \textbf{0.8136} & 100.0 & \underline{0.0773} & \underline{0.1195} & \underline{0.3598} & 100.0 \\
 & 192 & \textbf{0.0219} & \textbf{0.0808} & \textbf{0.7714} & 100.0 & \underline{0.0718} & \underline{0.1108} & \underline{0.3216} & 100.0 \\
 & 336 & \textbf{0.0232} & \textbf{0.0823} & \textbf{0.7242} & 100.0 & \underline{0.0669} & \underline{0.1062} & \underline{0.2901} & 99.3 \\
\midrule
\multirow{3}{*}{ETTh2} & 96 & \textbf{0.0305} & \textbf{0.1041} & \textbf{0.7302} & 100.0 & \underline{0.0688} & \underline{0.1327} & \underline{0.3859} & 99.8 \\
 & 192 & \textbf{0.0295} & \textbf{0.0999} & \textbf{0.6835} & 100.0 & \underline{0.0591} & \underline{0.1269} & \underline{0.3705} & 100.0 \\
 & 336 & \textbf{0.0292} & \textbf{0.0993} & \textbf{0.6456} & 100.0 & \underline{0.0531} & \underline{0.1186} & \underline{0.3596} & 100.0 \\
\midrule
\multirow{3}{*}{ETTm1} & 96 & \textbf{0.0161} & \textbf{0.0669} & \textbf{0.8713} & 100.0 & \underline{0.0411} & \underline{0.0947} & \underline{0.6852} & 100.0 \\
 & 192 & \textbf{0.0181} & \textbf{0.0685} & \textbf{0.8338} & 100.0 & \underline{0.0671} & \underline{0.0988} & \underline{0.4556} & 99.7 \\
 & 336 & \textbf{0.0276} & \textbf{0.0704} & \textbf{0.7341} & 100.0 & \underline{0.0719} & \underline{0.1014} & \underline{0.3305} & 99.7 \\
\midrule
\multirow{3}{*}{ETTm2} & 96 & \textbf{0.0279} & \textbf{0.0982} & \textbf{0.7944} & 100.0 & \underline{0.0514} & \underline{0.1240} & \underline{0.6143} & 100.0 \\
 & 192 & \textbf{0.0295} & \textbf{0.1016} & \textbf{0.7532} & 100.0 & \underline{0.0650} & \underline{0.1247} & \underline{0.4423} & 99.3 \\
 & 336 & \textbf{0.0332} & \textbf{0.0968} & \textbf{0.6689} & 100.0 & \underline{0.0657} & \underline{0.1213} & \underline{0.3367} & 100.0 \\
\midrule
\multirow{3}{*}{Weather} & 96 & \textbf{0.0159} & \textbf{0.0622} & \textbf{0.8599} & 100.0 & \underline{0.0354} & \underline{0.0839} & \underline{0.7228} & 99.9 \\
 & 192 & \textbf{0.0159} & \textbf{0.0592} & \textbf{0.8427} & 100.0 & \underline{0.0391} & \underline{0.0822} & \underline{0.6776} & 99.6 \\
 & 336 & \textbf{0.0213} & \textbf{0.0648} & \textbf{0.7701} & 100.0 & \underline{0.0553} & \underline{0.0861} & \underline{0.4938} & 99.7 \\
\midrule
\multirow{3}{*}{Electricity} & 96 & \textbf{0.0301} & \textbf{0.0923} & \textbf{0.8112} & 100.0 & \underline{0.1472} & \underline{0.1253} & \underline{0.2005} & 99.7 \\
 & 192 & \textbf{0.0284} & \textbf{0.0845} & \textbf{0.7982} & 99.8 & \underline{0.1480} & \underline{0.1066} & \underline{0.1049} & 99.6 \\
 & 336 & \textbf{0.0280} & \textbf{0.0824} & \textbf{0.7811} & 99.9 & \underline{0.1412} & \underline{0.0970} & \underline{0.0562} & 99.9 \\
\midrule
\multirow{3}{*}{Exchange} & 96 & \underline{0.0795} & \textbf{0.1643} & \textbf{0.4815} & 100.0 & \textbf{0.0788} & \underline{0.1861} & \underline{0.4406} & 100.0 \\
 & 192 & \underline{0.0837} & \textbf{0.1481} & \textbf{0.5180} & 100.0 & \textbf{0.0751} & \underline{0.1668} & \underline{0.4722} & 100.0 \\
 & 336 & \underline{0.0870} & \underline{0.1449} & \textbf{0.5159} & 100.0 & \textbf{0.0665} & \textbf{0.1385} & \underline{0.5109} & 100.0 \\
\midrule
\multirow{3}{*}{Web} & 96 & \underline{0.0326} & \underline{0.1162} & \underline{0.3672} & 100.0 & \textbf{0.0285} & \textbf{0.1017} & \textbf{0.4987} & 99.9 \\
 & 192 & \underline{0.0246} & \underline{0.1002} & \underline{0.3024} & 100.0 & \textbf{0.0191} & \textbf{0.0831} & \textbf{0.5365} & 99.9 \\
 & 336 & \underline{0.0175} & \underline{0.0823} & \underline{0.2664} & 99.9 & \textbf{0.0126} & \textbf{0.0661} & \textbf{0.5606} & 100.0 \\
\bottomrule
\end{tabular*}
\vspace{-0.6em}
\end{table*}

\textbf{Detailed Reward Design Ablation}
% \label{app:reward_ablation_details}
Table~\ref{tab:reward_ablation_detail} reports the detailed reward design ablation results for each benchmark dataset and generation length. Compared with removing the statistical reward or removing both correlation and statistical rewards, the full reward generally provides better overall generation quality, especially in Pearson correlation. This indicates that the correlation and statistical terms are important for preserving temporal shape and avoiding structurally weak outputs. In several settings such as Web, removing these terms can reduce MSE or DTW, but it also sharply decreases Pearson correlation, suggesting that lower local error does not necessarily imply better preservation of temporal patterns. These results support the use of a balanced reward design that jointly considers pointwise accuracy, temporal alignment, and statistical consistency.

\begin{table*}[t]
\centering
\setlength{\abovecaptionskip}{2pt}
\setlength{\belowcaptionskip}{1pt}
\caption{Reward-design ablation; SR is pass@1 success rate.}
\label{tab:reward_ablation_detail}
\scriptsize
\setlength{\tabcolsep}{3.0pt}
\renewcommand{\arraystretch}{0.80}
\begin{tabular*}{\textwidth}{@{\extracolsep{\fill}}ll|cccc|cccc|cccc@{}}
\toprule
 & & \multicolumn{4}{c|}{Full} & \multicolumn{4}{c|}{w/o stat} & \multicolumn{4}{c}{w/o corr, stat} \\
\cmidrule(lr){3-14}
Datasets & Length & MSE$\downarrow$ & DTW$\downarrow$ & Pearson$\uparrow$ & SR(\%)$\uparrow$ & MSE$\downarrow$ & DTW$\downarrow$ & Pearson$\uparrow$ & SR(\%)$\uparrow$ & MSE$\downarrow$ & DTW$\downarrow$ & Pearson$\uparrow$ & SR(\%)$\uparrow$ \\
\midrule
\multirow{3}{*}{ETTh1} & 96 & \textbf{0.0202} & \textbf{0.0799} & \textbf{0.8136} & 100.0 & \underline{0.0391} & \underline{0.1091} & \underline{0.6418} & 100.0 & 0.0400 & 0.1184 & 0.5951 & 100.0 \\
 & 192 & \textbf{0.0219} & \textbf{0.0808} & \textbf{0.7714} & 100.0 & 0.0395 & \underline{0.1080} & \underline{0.5899} & 100.0 & \underline{0.0393} & 0.1140 & 0.5331 & 100.0 \\
 & 336 & \textbf{0.0232} & \textbf{0.0823} & \textbf{0.7242} & 100.0 & 0.0391 & \underline{0.1085} & \underline{0.5377} & 100.0 & \underline{0.0379} & 0.1129 & 0.4720 & 100.0 \\
\midrule
\multirow{3}{*}{ETTh2} & 96 & \textbf{0.0305} & \textbf{0.1041} & \textbf{0.7302} & 100.0 & 0.0552 & \underline{0.1268} & \underline{0.5055} & 100.0 & \underline{0.0505} & 0.1327 & 0.4626 & 100.0 \\
 & 192 & \textbf{0.0295} & \textbf{0.0999} & \textbf{0.6835} & 100.0 & 0.0517 & \underline{0.1238} & \underline{0.4476} & 100.0 & \underline{0.0471} & 0.1272 & 0.3860 & 100.0 \\
 & 336 & \textbf{0.0292} & \textbf{0.0993} & \textbf{0.6456} & 100.0 & 0.0481 & \underline{0.1211} & \underline{0.4188} & 100.0 & \underline{0.0439} & 0.1213 & 0.3477 & 100.0 \\
\midrule
\multirow{3}{*}{ETTm1} & 96 & \textbf{0.0161} & \textbf{0.0669} & \textbf{0.8713} & 100.0 & \underline{0.0366} & \underline{0.1040} & \underline{0.7126} & 100.0 & 0.0494 & 0.1310 & 0.5885 & 100.0 \\
 & 192 & \textbf{0.0181} & \textbf{0.0685} & \textbf{0.8338} & 100.0 & \underline{0.0365} & \underline{0.1049} & \underline{0.6660} & 100.0 & 0.0440 & 0.1152 & 0.5589 & 100.0 \\
 & 336 & \textbf{0.0276} & \textbf{0.0704} & \textbf{0.7341} & 100.0 & 0.0520 & 0.1261 & \underline{0.5008} & 100.0 & \underline{0.0464} & \underline{0.1230} & 0.4419 & 100.0 \\
\midrule
\multirow{3}{*}{ETTm2} & 96 & \textbf{0.0279} & \textbf{0.0982} & \textbf{0.7944} & 100.0 & \underline{0.0604} & \underline{0.1329} & \underline{0.5461} & 100.0 & 0.0606 & 0.1512 & 0.4748 & 100.0 \\
 & 192 & \textbf{0.0295} & \textbf{0.1016} & \textbf{0.7532} & 100.0 & 0.0563 & \underline{0.1278} & \underline{0.5076} & 100.0 & \underline{0.0546} & 0.1395 & 0.4192 & 100.0 \\
 & 336 & \textbf{0.0332} & \textbf{0.0968} & \textbf{0.6689} & 100.0 & 0.0571 & \underline{0.1322} & \underline{0.4210} & 100.0 & \underline{0.0520} & 0.1359 & 0.3559 & 100.0 \\
\midrule
\multirow{3}{*}{Weather} & 96 & \textbf{0.0159} & \textbf{0.0622} & \textbf{0.8599} & 100.0 & \underline{0.0457} & \underline{0.1129} & \underline{0.6469} & 100.0 & 0.0538 & 0.1423 & 0.5986 & 100.0 \\
 & 192 & \textbf{0.0159} & \textbf{0.0592} & \textbf{0.8427} & 100.0 & \underline{0.0472} & \underline{0.1135} & \underline{0.6049} & 100.0 & 0.0556 & 0.1387 & 0.4855 & 99.9 \\
 & 336 & \textbf{0.0213} & \textbf{0.0648} & \textbf{0.7701} & 100.0 & 0.0464 & \underline{0.1182} & \underline{0.5470} & 100.0 & \underline{0.0447} & 0.1226 & 0.5038 & 100.0 \\
\midrule
\multirow{3}{*}{Electricity} & 96 & \textbf{0.0301} & \textbf{0.0923} & \textbf{0.8112} & 100.0 & 0.0520 & \underline{0.1223} & \underline{0.6827} & 99.9 & \underline{0.0485} & 0.1245 & 0.6534 & 100.0 \\
 & 192 & \textbf{0.0284} & \textbf{0.0845} & \textbf{0.7982} & 99.8 & 0.0482 & \underline{0.1179} & \underline{0.6678} & 99.8 & \underline{0.0448} & 0.1186 & 0.6341 & 100.0 \\
 & 336 & \textbf{0.0280} & \textbf{0.0824} & \textbf{0.7811} & 99.9 & 0.0462 & 0.1157 & \underline{0.6494} & 99.9 & \underline{0.0419} & \underline{0.1148} & 0.6178 & 99.9 \\
\midrule
\multirow{3}{*}{Exchange} & 96 & 0.0795 & 0.1643 & \textbf{0.4815} & 100.0 & \underline{0.0740} & \textbf{0.1554} & 0.4565 & 100.0 & \textbf{0.0603} & \underline{0.1598} & \underline{0.4717} & 100.0 \\
 & 192 & 0.0837 & \underline{0.1481} & \textbf{0.5180} & 100.0 & \underline{0.0685} & \textbf{0.1389} & \underline{0.5039} & 100.0 & \textbf{0.0599} & 0.1597 & 0.4961 & 98.9 \\
 & 336 & 0.0870 & \underline{0.1449} & \textbf{0.5159} & 100.0 & \textbf{0.0574} & \textbf{0.1253} & \underline{0.4990} & 100.0 & \underline{0.0598} & 0.1591 & 0.4875 & 100.0 \\
\midrule
\multirow{3}{*}{Web} & 96 & \underline{0.0326} & 0.1162 & \textbf{0.3672} & 100.0 & 0.0456 & \underline{0.1140} & \underline{0.1396} & 99.9 & \textbf{0.0317} & \textbf{0.0869} & 0.1121 & 100.0 \\
 & 192 & \underline{0.0246} & 0.1002 & \textbf{0.3024} & 100.0 & 0.0334 & \underline{0.0974} & \underline{0.0841} & 98.2 & \textbf{0.0220} & \textbf{0.0705} & 0.0784 & 100.0 \\
 & 336 & \underline{0.0175} & \underline{0.0823} & \textbf{0.2664} & 99.9 & 0.0232 & 0.0825 & \underline{0.0544} & 99.0 & \textbf{0.0154} & \textbf{0.0588} & 0.0505 & 100.0 \\
\bottomrule
\end{tabular*}
\vspace{-0.6em}
\end{table*}

\section{Detailed Template Robustness Analysis}
\label{app:template_robustness_details}

Table~\ref{tab:template_robustness_detail} reports the detailed template-dependence results for each benchmark dataset and generation length. The perturbed descriptions preserve the same temporal attributes as the original descriptions, but change the wording and attribute order. Compared with the SFT-warmup model under perturbed descriptions, the final CodeTS model achieves substantially better MSE, DTW, Pearson correlation, and pass@1 success rate in most settings. Although performance still decreases compared with the original descriptions, the smaller degradation shows that RLVR improves robustness to description perturbations and reduces reliance on fixed training templates.

\begin{table*}[t]
\centering
\setlength{\abovecaptionskip}{2pt}
\setlength{\belowcaptionskip}{1pt}
\caption{Template-dependence ablation; SR is pass@1 success rate.}
\label{tab:template_robustness_detail}
\scriptsize
\setlength{\tabcolsep}{3.0pt}
\renewcommand{\arraystretch}{0.80}
\begin{tabular*}{\textwidth}{@{\extracolsep{\fill}}ll|cccc|cccc|cccc@{}}
\toprule
 & & \multicolumn{4}{c|}{Full} & \multicolumn{4}{c|}{Full w/ perturbation} & \multicolumn{4}{c}{SFT w/ perturbation} \\
\cmidrule(lr){3-14}
Datasets & Length & MSE$\downarrow$ & DTW$\downarrow$ & Pearson$\uparrow$ & SR(\%)$\uparrow$ & MSE$\downarrow$ & DTW$\downarrow$ & Pearson$\uparrow$ & SR(\%)$\uparrow$ & MSE$\downarrow$ & DTW$\downarrow$ & Pearson$\uparrow$ & SR(\%)$\uparrow$ \\
\midrule
\multirow{3}{*}{ETTh1} & 96 & \textbf{0.0202} & \textbf{0.0799} & \textbf{0.8136} & 100.0 & \underline{0.0440} & \underline{0.1073} & \underline{0.6578} & 92.4 & 0.0997 & 0.1481 & 0.1766 & 47.0 \\
 & 192 & \textbf{0.0219} & \textbf{0.0808} & \textbf{0.7714} & 100.0 & \underline{0.0426} & \underline{0.1104} & \underline{0.5815} & 93.7 & 0.1005 & 0.1483 & 0.1307 & 46.0 \\
 & 336 & \textbf{0.0232} & \textbf{0.0823} & \textbf{0.7242} & 100.0 & \underline{0.0426} & \underline{0.1155} & \underline{0.5138} & 90.5 & 0.0836 & 0.1359 & 0.1310 & 38.1 \\
\midrule
\multirow{3}{*}{ETTh2} & 96 & \textbf{0.0305} & \textbf{0.1041} & \textbf{0.7302} & 100.0 & \underline{0.0545} & \underline{0.1275} & \underline{0.5432} & 92.4 & 0.1045 & 0.1635 & 0.1782 & 44.2 \\
 & 192 & \textbf{0.0295} & \textbf{0.0999} & \textbf{0.6835} & 100.0 & \underline{0.0481} & \underline{0.1235} & \underline{0.4967} & 95.6 & 0.0832 & 0.1477 & 0.1710 & 51.2 \\
 & 336 & \textbf{0.0292} & \textbf{0.0993} & \textbf{0.6456} & 100.0 & \underline{0.0442} & \underline{0.1182} & \underline{0.4600} & 94.6 & 0.0747 & 0.1491 & 0.1945 & 43.5 \\
\midrule
\multirow{3}{*}{ETTm1} & 96 & \textbf{0.0161} & \textbf{0.0669} & \textbf{0.8713} & 100.0 & \underline{0.0329} & \underline{0.0932} & \underline{0.7589} & 93.2 & 0.0739 & 0.1362 & 0.4775 & 46.7 \\
 & 192 & \textbf{0.0181} & \textbf{0.0685} & \textbf{0.8338} & 100.0 & \underline{0.0386} & \underline{0.1005} & \underline{0.6783} & 93.9 & 0.0946 & 0.1440 & 0.2986 & 47.0 \\
 & 336 & \textbf{0.0276} & \textbf{0.0704} & \textbf{0.7341} & 100.0 & \underline{0.0563} & \underline{0.1155} & \underline{0.5131} & 94.0 & 0.0905 & 0.1502 & 0.1617 & 47.5 \\
\midrule
\multirow{3}{*}{ETTm2} & 96 & \textbf{0.0279} & \textbf{0.0982} & \textbf{0.7944} & 100.0 & \underline{0.0528} & \underline{0.1236} & \underline{0.6659} & 93.9 & 0.0868 & 0.1529 & 0.3809 & 48.2 \\
 & 192 & \textbf{0.0295} & \textbf{0.1016} & \textbf{0.7532} & 100.0 & \underline{0.0524} & \underline{0.1260} & \underline{0.5913} & 93.5 & 0.1045 & 0.1566 & 0.2562 & 51.2 \\
 & 336 & \textbf{0.0332} & \textbf{0.0968} & \textbf{0.6689} & 100.0 & \underline{0.0603} & \underline{0.1289} & \underline{0.4786} & 93.3 & 0.0832 & 0.1511 & 0.1622 & 49.4 \\
\midrule
\multirow{3}{*}{Weather} & 96 & \textbf{0.0159} & \textbf{0.0622} & \textbf{0.8599} & 100.0 & \underline{0.0298} & \underline{0.0879} & \underline{0.7644} & 94.0 & 0.0768 & 0.1355 & 0.4844 & 47.5 \\
 & 192 & \textbf{0.0159} & \textbf{0.0592} & \textbf{0.8427} & 100.0 & \underline{0.0290} & \underline{0.0859} & \underline{0.7451} & 92.9 & 0.0704 & 0.1285 & 0.4409 & 45.8 \\
 & 336 & \textbf{0.0213} & \textbf{0.0648} & \textbf{0.7701} & 100.0 & \underline{0.0373} & \underline{0.0983} & \underline{0.6353} & 93.8 & 0.0773 & 0.1369 & 0.3022 & 44.5 \\
\midrule
\multirow{3}{*}{Electricity} & 96 & \textbf{0.0301} & \textbf{0.0923} & \textbf{0.8112} & 100.0 & \underline{0.0588} & \underline{0.1249} & \underline{0.6690} & 94.0 & 0.1700 & 0.1761 & 0.1197 & 38.7 \\
 & 192 & \textbf{0.0284} & \textbf{0.0845} & \textbf{0.7982} & 99.8 & \underline{0.0641} & \underline{0.1255} & \underline{0.6102} & 91.3 & 0.1663 & 0.1626 & 0.0547 & 37.0 \\
 & 336 & \textbf{0.0280} & \textbf{0.0824} & \textbf{0.7811} & 99.9 & \underline{0.0604} & \underline{0.1240} & \underline{0.5693} & 91.4 & 0.1529 & 0.1597 & 0.0340 & 38.8 \\
\midrule
\multirow{3}{*}{Exchange} & 96 & \textbf{0.0795} & \textbf{0.1643} & \textbf{0.4815} & 100.0 & \underline{0.1000} & \underline{0.1778} & \underline{0.4082} & 92.3 & 0.1672 & 0.2177 & 0.1785 & 49.5 \\
 & 192 & \textbf{0.0837} & \textbf{0.1481} & \textbf{0.5180} & 100.0 & \underline{0.0866} & \underline{0.1546} & \underline{0.4058} & 95.6 & 0.1067 & 0.1579 & 0.2739 & 42.9 \\
 & 336 & 0.0870 & 0.1449 & \textbf{0.5159} & 100.0 & \textbf{0.0808} & \underline{0.1397} & \underline{0.4756} & 92.9 & \underline{0.0813} & \textbf{0.1304} & 0.2678 & 53.6 \\
\midrule
\multirow{3}{*}{Web} & 96 & \textbf{0.0326} & \textbf{0.1162} & \textbf{0.3672} & 100.0 & \underline{0.0370} & 0.1178 & \underline{0.3033} & 96.5 & 0.0481 & \underline{0.1171} & 0.1015 & 30.3 \\
 & 192 & \textbf{0.0246} & \underline{0.1002} & \textbf{0.3024} & 100.0 & \underline{0.0271} & 0.1012 & \underline{0.2601} & 96.5 & 0.0349 & \textbf{0.1000} & 0.0795 & 62.5 \\
 & 336 & \textbf{0.0175} & \textbf{0.0823} & \textbf{0.2664} & 99.9 & \underline{0.0197} & \underline{0.0843} & \underline{0.2276} & 96.1 & 0.0245 & 0.0854 & 0.0579 & 64.8 \\
\bottomrule
\end{tabular*}
\vspace{-0.6em}
\end{table*}

\section{Reinforcement Learning Training Dynamics and Effectiveness Analysis}
\label{app:rl_dynamics}

\begin{figure}[t]
\centering
\includegraphics[width=\textwidth]{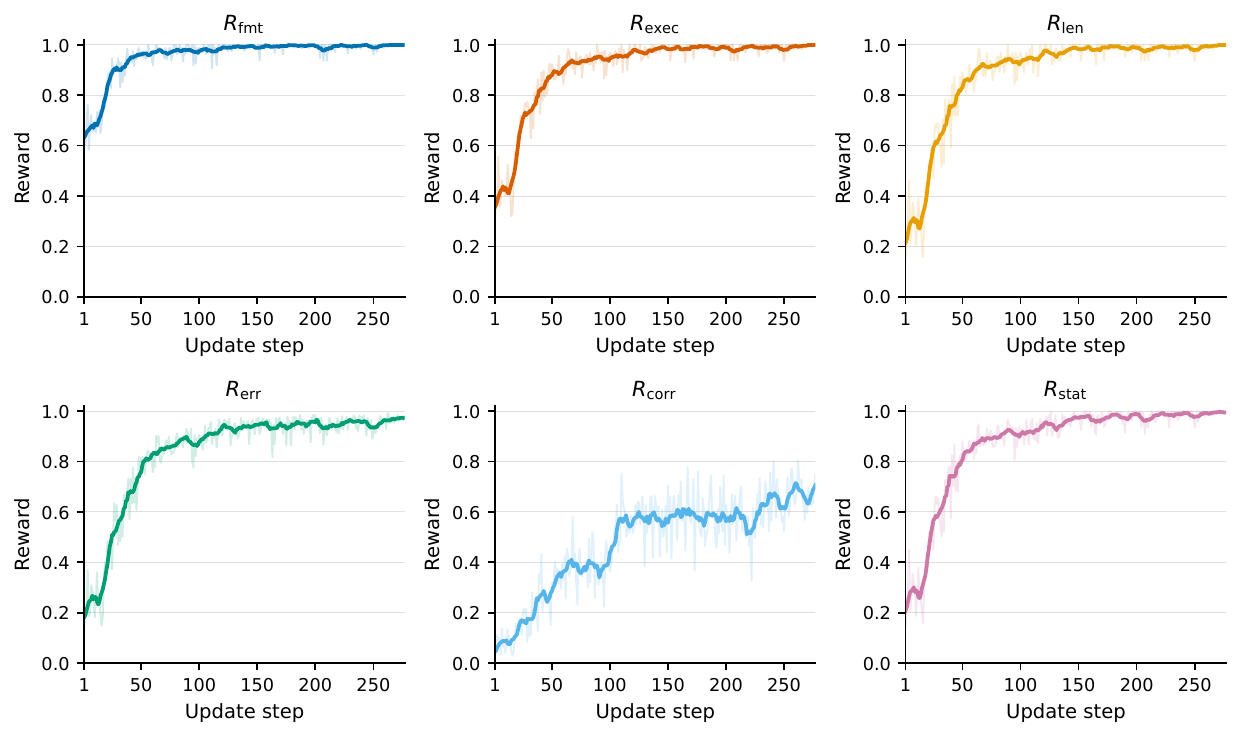}
\caption{Training dynamics of the six RLVR reward functions.}
\label{fig:app_training_dynamics_grid}
\end{figure}

During RLVR, we track six reward functions: \(R_{\mathrm{fmt}}\),
\(R_{\mathrm{exec}}\), \(R_{\mathrm{len}}\), \(R_{\mathrm{err}}\),
\(R_{\mathrm{corr}}\), and \(R_{\mathrm{stat}}\).
Figure~\ref{fig:app_training_dynamics_grid} shows their training trajectories.

The first three scores rise quickly because SFT already teaches the JSON schema, runnable scaffold, and explicit length parameter. 
The time-series terms improve more gradually, reflecting the harder search for code that matches target values, shape, and variability.

This trend is consistent with the reward-ablation results in the main paper. 
Pointwise error alone can produce numerically plausible but structurally weak sequences; shape correlation improves global alignment, and variance preservation discourages low-variance collapse.

\section{Prompts for Training and Inference}
\label{app:prompts}

This section documents the prompts used by CodeTS and the zero-shot baselines. 
For SFT, RLVR, and fine-tuned model inference, we use the same prompt skeleton so that the output interface remains consistent across training and evaluation. The system message defines the model role, and the user message asks the model to translate a natural language time series description into a normalized JSON object containing \texttt{params} and \texttt{code}. During SFT, the reference JSON object is used as the supervised target. During RLVR, the same prompt is used to sample candidate code, while the paired time series is used only for execution-based reward computation. During inference, the test-time description is appended to the same user instruction.

\begin{figure}[!htbp]
\centering
\includegraphics[width=\linewidth,trim=18pt 5pt 19pt 5pt,clip]{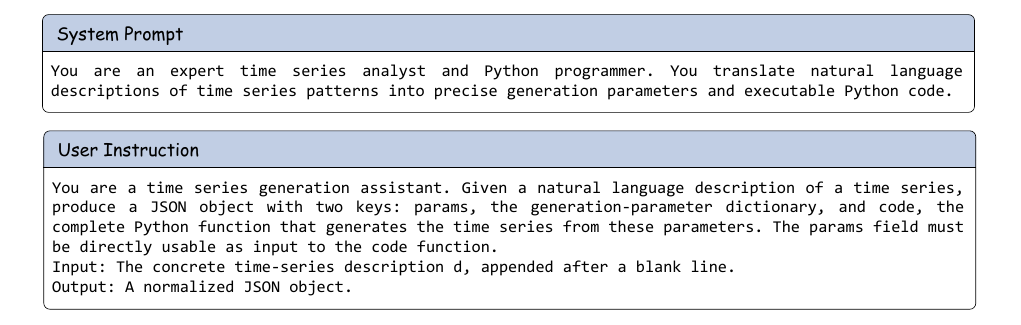}
\caption{Prompt skeleton used for SFT, RLVR, and fine-tuned CodeTS
inference. The concrete time-series description is appended to the user
message; the target JSON object is provided only in SFT, while RLVR uses the
paired time series only for reward computation.}
\label{fig:fintuning_prompt}
\end{figure}

For zero-shot LLM baselines, we use a separate and more explicit prompt because these models have not been task-specifically tuned to follow the normalized CodeTS output interface. 
The zero-shot prompt therefore includes the required JSON schema, the required \texttt{generate\_ts(params)} function signature, the available Python libraries, safety constraints, and an example output format. 
In this setting, the user message contains only the test description, while the system prompt provides the task definition and formatting constraints.

\begin{figure}[!htbp]
\centering
\includegraphics[width=\linewidth,trim=18pt 25pt 19pt 23pt,clip]{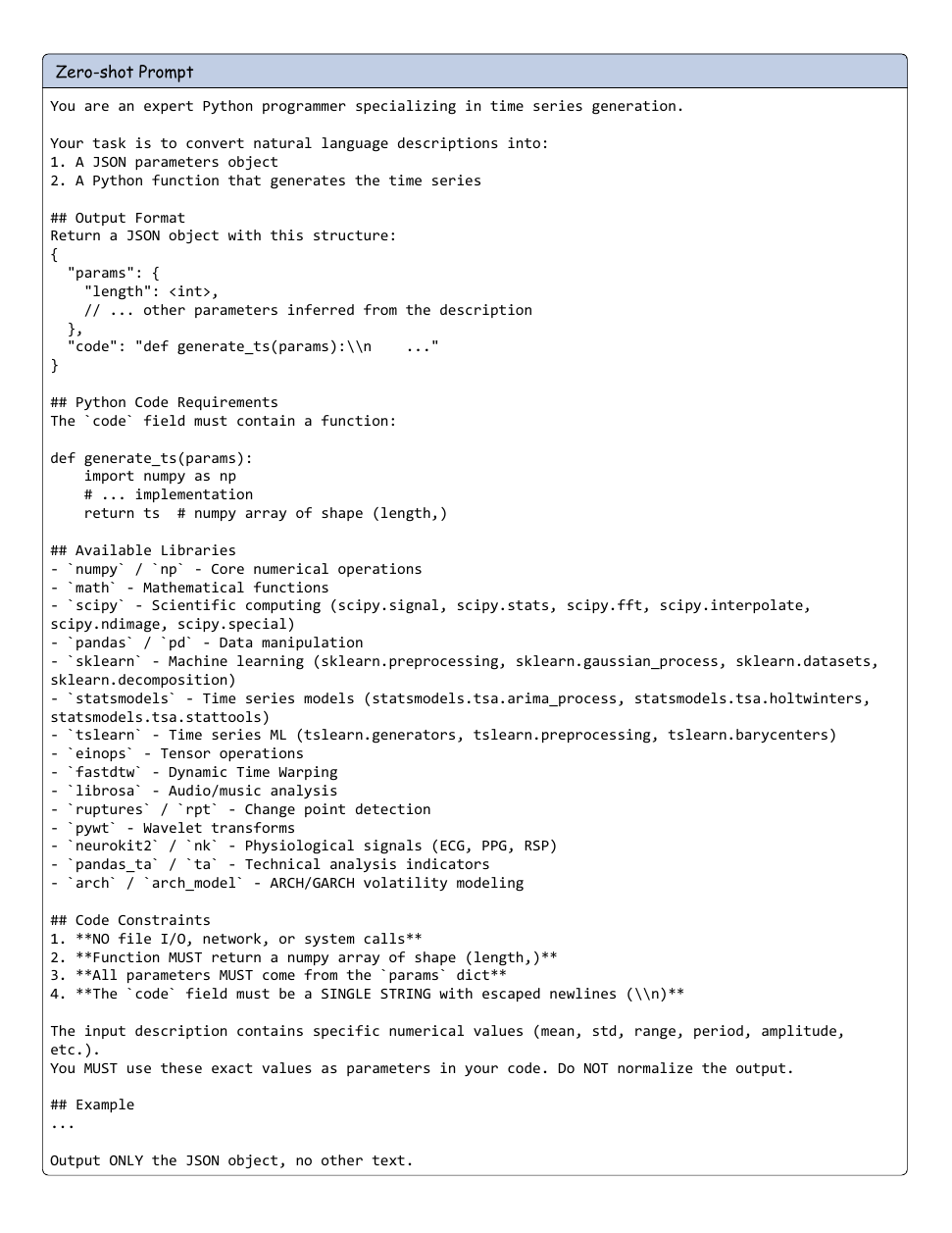}
\caption{Prompt used for zero-shot LLM baselines. Compared with the
fine-tuned-model prompt, this prompt is more constrained and includes explicit
format, code, library, and safety requirements to improve parseability and
executability without task-specific fine-tuning.}
\label{fig:zero_shot_prompt}
\end{figure}
% \FloatBarrier

\section{Broader Impact}
\label{app:broader_impact}

Our work introduces CodeTS, a verifiable Text-to-Code-to-TS generation framework that employs executable code as an explicit intermediate interface between natural language temporal descriptions and synthetic time series generation. 
This design has the potential to facilitate time series modeling in data-scarce settings, including healthcare, energy systems, transportation, finance, and climate analysis, where acquiring representative observations may be costly, privacy-sensitive, or constrained by the rarity of relevant events. 
By rendering the generation process executable and inspectable, CodeTS further enables users to construct targeted temporal scenarios, stress-test downstream models, and examine whether generated samples conform to the specified temporal requirements.

Nevertheless, synthetic time series should not be regarded as a replacement for validated domain observations in high-stakes applications.
Ambiguous or underspecified descriptions, biases inherited from training data or reference programs, and excessive reliance on generated samples may result in misleading evaluations or undesirable downstream model behavior. 
In addition, the use of executable code as an intermediate representation introduces security considerations, since candidate programs must be evaluated through execution. 
In this work, normalized code, format validation, sandboxed execution, and success-rate reporting serve as partial safeguards. 
However, real-world deployment should further incorporate domain-specific validation, provenance tracking, privacy assessment, and explicit disclosure whenever synthetic time series are used for analysis, evaluation, or model development.

\section{Additional Case Studies}
\label{app:case_studies}

We provide three representative successful cases selected from CodeTS evaluation outputs, prioritizing local-event richness rather than only the lowest global error. 
Each case forms a Text-Code-TS triplet: the natural language description \(d\), the normalized generated program \(y\) containing \texttt{params} and \texttt{code}, and the executed time series \(\hat{x}=\mathrm{Execute}(y)\). 
Each case starts on a separate page so that the compressed text description, time-series visualization, folded parameter summary, execution metrics, and unframed executable code can be inspected together.

% Case-study appendix block from results/code4ts/grpo.
\clearpage
\subsection{Case 1: Electricity-336 Client 314}
\label{app:case_study_1}
\vspace{-0.8em}
{\fontsize{5.65}{4.9}\selectfont
\noindent\textbf{Text description.} This sequence is 314. The time series spans 336 points with mean 8271.58, std 4722.22, range [2115.00, 16956.00]. Flat linear trend from baseline 7788.97 to 8754.19 from position 0 to 335. Pure sinusoidal seasonality with period 24.0, amplitude 6468.78, phase 1.96 rad, 14 cycles, constant envelope. One smooth+up from position 19 to 25 (peak at 22), magnitude 0.24. One spike+down from position 23 to 25 (peak at 24), magnitude 0.19. One spike+up from position 27 to 31 (peak at 29), magnitude 0.17. One spike+down from position 33 to 35 (peak at 34), magnitude 0.17. One spike+up from position 44 to 46 (peak at 45), magnitude 0.19. One smooth+down from position 47 to 51 (peak at 49), magnitude 0.22. One spike+up from position 68 to 70 (peak at 69), magnitude 0.22. One spike+down from position 71 to 75 (peak at 73), magnitude 0.24. One smooth+up from position 91 to 95 (peak at 93), magnitude 0.24. One spike+down from position 96 to 98 (peak at 97), magnitude 0.19. One smooth+up from position 99 to 103 (peak at 101), magnitude 0.19. One smooth+down from position 118 to 124 (peak at 121), magnitude 0.28. One spike+up from position 130 to 134 (peak at 132), magnitude 0.28. One spike+down from position 134 to 136 (peak at 135), magnitude 0.29. One smooth+up from position 142 to 156 (peak at 149), magnitude 0.29. One smooth+down from position 154 to 164 (peak at 159), magnitude 0.38. One smooth+up from position 165 to 181 (peak at 173), magnitude 0.29. One smooth+down from position 181 to 185 (peak at 183), magnitude 0.29. One smooth+up from position 188 to 192 (peak at 190), magnitude 0.16. One spike+down from position 192 to 194 (peak at 193), magnitude 0.16. One spike+up from position 201 to 205 (peak at 203), magnitude 0.26. One spike+down from position 207 to 209 (peak at 208), magnitude 0.17. One spike+up from position 212 to 214 (peak at 213), magnitude 0.17. One spike+down from position 216 to 218 (peak at 217), magnitude 0.21. One spike+up from position 220 to 222 (peak at 221), magnitude 0.15. One spike+down from position 225 to 227 (peak at 226), magnitude 0.15. One smooth+up from position 231 to 243 (peak at 237), magnitude 0.21. One spike+down from position 240 to 242 (peak at 241), magnitude 0.20. One spike+up from position 243 to 247 (peak at 245), magnitude 0.20. One spike+down from position 249 to 251 (peak at 250), magnitude 0.17. One spike+up from position 252 to 254 (peak at 253), magnitude 0.17. One spike+down from position 264 to 266 (peak at 265), magnitude 0.20. One spike+up from position 267 to 271 (peak at 269), magnitude 0.15. One spike+down from position 273 to 275 (peak at 274), magnitude 0.15. One spike+up from position 276 to 278 (peak at 277), magnitude 0.20. One spike+down from position 287 to 291 (peak at 289), magnitude 0.26. One spike+up from position 299 to 301 (peak at 300), magnitude 0.26. One spike+down from position 302 to 304 (peak at 303), magnitude 0.16. One smooth+up from position 308 to 312 (peak at 310), magnitude 0.16. One spike+down from position 311 to 315 (peak at 313), magnitude 0.26. One spike+up from position 323 to 325 (peak at 324), magnitude 0.17. One spike+down from position 326 to 328 (peak at 327), magnitude 0.17. Noise std\_deviation 1309.070.\par}
{\fontsize{6.0}{5.6}\selectfont
\noindent\textbf{Execution.} \(L=336\); 42 events; MSE 0.0063; DTW 0.0453; PC 0.9632.\par}
\vspace{-0.4em}
\begin{center}
\includegraphics[width=0.82\linewidth,height=0.105\textheight,trim=4pt 8pt 4pt 8pt,clip,keepaspectratio]{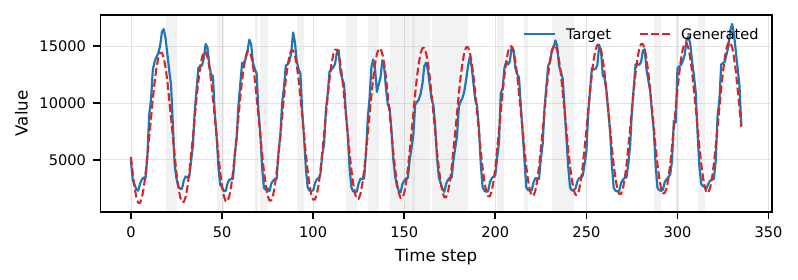}
\end{center}
\vspace{-1.5em}
\noindent\textbf{Folded params.}
\begin{lstlisting}[basicstyle=\ttfamily\fontsize{4.9}{5.1}\selectfont]
length=336; target_mean=8271.584; target_std=4722.224; noise_std=1309.07
trend: linear(0-335): 7788.974 -> 8754.194
seasonality: sinusoidal, period=24.0, amplitude=6468.784, phase=1.962, envelope=constant
events: 42 total = 30 spike + 12 smooth; 22 with magnitude >= 0.2; max magnitude=0.384
top events: smooth+down at 159 [154,164], mag=0.384; spike+down at 135 [134,136], mag=0.294; 
smooth+up at 149 [142,156], mag=0.294; smooth+up at 173 [165,181], mag=0.294
\end{lstlisting}
\vspace{-0.8em}
\noindent\textbf{Generated code.}
\begin{lstlisting}[language=Python,basicstyle=\fontsize{5.9}{5.1}\ttfamily]
def generate_ts(params):
    import numpy as np
    length = params['length']
    t = np.arange(length, dtype=float)
    def _trend_seg(n, sv, ev, stype, extra):
        if n <= 0:
            return np.array([])
        if n == 1:
            return np.array([sv])
        tn = np.arange(n, dtype=float) / (n - 1)
        return sv + (ev - sv) * tn
    trend = np.zeros(length)
    for _s in params['trend_segments']:
        _a, _b = (_s.get('start', 0), _s.get('end', length - 1))
        trend[_a:_b + 1] = _trend_seg(_b - _a + 1, _s.get('start_value', 0.0), 
        _s.get('end_value', _s.get('start_value', 0.0)), _s.get('trend_type', 'constant'), _s)
    _last = params['trend_segments'][-1]
    _last_end = _last.get('end', length - 1)
    if _last_end < length - 1:
        trend[_last_end + 1:] = trend[_last_end]
    seasonality_type = params.get('seasonality_type', None)
    seasonality = np.zeros(length)
    def _build_seasonal(t_arr, s_type, period, amplitude, phase, harmonics, envelope_type, length_seg):
        seg = np.zeros(len(t_arr))
        if s_type is None or s_type == 'none' or amplitude == 0:
            return seg
        t_local = np.arange(len(t_arr), dtype=float)
        envelope = np.full(len(t_arr), 1.0)
        omega = 2.0 * np.pi * t_arr / period + phase
        seg = amplitude * np.cos(omega)
        for h in harmonics:
            order = h.get('order', 2)
            amp_ratio = h.get('amplitude_ratio', h.get('relative_amplitude', 0.0))
            seg += amplitude * amp_ratio * np.cos(order * 2.0 * np.pi * t_arr / period + phase)
        return seg * envelope
    seasonal_regions = params.get('seasonal_regions', None)
    period = params.get('period', length)
    amplitude = params.get('amplitude', 0.0)
    phase = params.get('phase', 0.0)
    harmonics = params.get('harmonics', [])
    envelope_type = params.get('amplitude_envelope_type', 'constant')
    seasonality = _build_seasonal(t, seasonality_type, period, amplitude, phase, harmonics, envelope_type, length)
    changes_array = np.zeros(length)
    for change in params.get('changes', []):
        ch_type = change.get('change_type', 'spike')
        direction = change.get('direction', 'up')
        tp = change.get('turning_point', length // 2)
        mag = abs(change.get('magnitude', 0.0))
        if direction == 'down':
            mag = -mag
        sp = change.get('start_position', None)
        ep = change.get('end_position', None)
        sigma = max(1.0, (ep - sp) / 4.0)
        changes_array += mag * np.exp(-0.5 * ((t - tp) / sigma) ** 2)
    raw_ts = trend + seasonality + changes_array
    raw_mean = params.get('target_mean', None).mean()
    raw_std = params.get('target_std', None).std()
    ts = (raw_ts - raw_mean) / raw_std * target_std + target_mean
    return ts
\end{lstlisting}

\clearpage
\subsection{Case 2: Weather-96 VPact}
\label{app:case_study_2}
\vspace{-0.8em}
{\fontsize{5.85}{5.1}\selectfont
\noindent\textbf{Text description.} This sequence is VPact (mbar). The time series spans 96 points with mean 10.28, std 0.36, range [9.43, 10.92]. Downward linear trend from baseline 10.74 to 9.72 from position 0 to 47. Upward linear trend from baseline 9.83 to 10.85 from position 48 to 95. Pure sinusoidal seasonality with period 48.0, amplitude 0.20, phase -3.13 rad, 2 cycles, increasing envelope. One spike+down from position 7 to 11 (peak at 9), magnitude 0.11. One spike+up from position 10 to 12 (peak at 11), magnitude 0.11. One spike+down from position 14 to 18 (peak at 16), magnitude 0.12. One spike+up from position 17 to 19 (peak at 18), magnitude 0.12. One spike+down from position 24 to 26 (peak at 25), magnitude 0.09. One spike+up from position 26 to 28 (peak at 27), magnitude 0.09. One smooth+down from position 24 to 34 (peak at 29), magnitude 0.26. One spike+up from position 30 to 34 (peak at 32), magnitude 0.26. One spike+down from position 34 to 36 (peak at 35), magnitude 0.10. One spike+up from position 35 to 37 (peak at 36), magnitude 0.10. One smooth+down from position 43 to 53 (peak at 48), magnitude 0.23. One smooth+up from position 61 to 67 (peak at 64), magnitude 0.23. One spike+down from position 68 to 70 (peak at 69), magnitude 0.28. One spike+up from position 72 to 74 (peak at 73), magnitude 0.36. One spike+down from position 73 to 77 (peak at 75), magnitude 0.36. One spike+up from position 78 to 80 (peak at 79), magnitude 0.14. One spike+down from position 81 to 83 (peak at 82), magnitude 0.14. One spike+up from position 83 to 85 (peak at 84), magnitude 0.34. One smooth+down from position 85 to 89 (peak at 87), magnitude 0.17. One spike+up from position 91 to 93 (peak at 92), magnitude 0.17. Noise std\_deviation 0.117.\par}
{\fontsize{6.3}{5.6}\selectfont
\noindent\textbf{Execution.} \(L=96\); 20 events; MSE 0.0027; DTW 0.0317; PC 0.9767.\par}
\vspace{-0.2em}
\begin{center}
\includegraphics[width=0.82\linewidth,height=0.105\textheight,trim=4pt 8pt 4pt 8pt,clip,keepaspectratio]{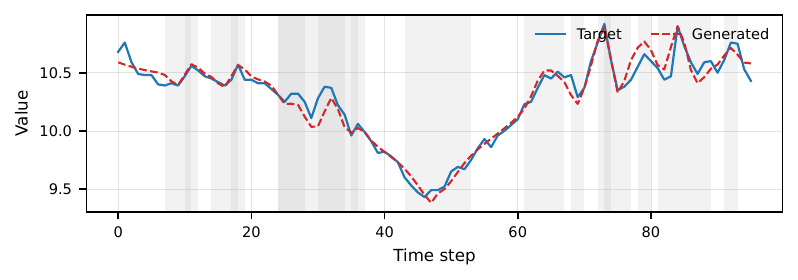}
\end{center}
\vspace{-1.5em}
\noindent\textbf{Folded params.}
\begin{lstlisting}[basicstyle=\ttfamily\fontsize{5.9}{6.1}\selectfont]
length=96; target_mean=10.283; target_std=0.3625; noise_std=0.1174
trend: linear(0-47): 10.744 -> 9.724; linear(48-95): 9.832 -> 10.852
seasonality: sinusoidal, period=48.0, amplitude=0.203, phase=-3.134, envelope=increasing
events: 20 total = 16 spike + 4 smooth; 8 with magnitude >= 0.2; max magnitude=0.364
top events: spike+up at 73 [72,74], mag=0.364; spike+down at 75 [73,77], mag=0.364; 
spike+up at 84 [83,85], mag=0.344; spike+down at 69 [68,70], mag=0.284
\end{lstlisting}
\vspace{-0.4em}
\noindent\textbf{Generated code.}
\begin{lstlisting}[language=Python,basicstyle=\fontsize{5.9}{6.1}\ttfamily]
def generate_ts(params):
    import numpy as np
    length = params['length']
    t = np.arange(length, dtype=float)
    def _trend_seg(n, sv, ev, stype, extra):
        if n <= 0:
            return np.array([])
        if n == 1:
            return np.array([sv])
        tn = np.arange(n, dtype=float) / (n - 1)
        return sv + (ev - sv) * tn
    trend = np.zeros(length)
    for _s in params['trend_segments']:
        _a, _b = (_s.get('start', 0), _s.get('end', length - 1))
        trend[_a:_b + 1] = _trend_seg(_b - _a + 1, _s.get('start_value', 0.0), 
        _s.get('end_value', _s.get('start_value', 0.0)), _s.get('trend_type', 'constant'), _s)
    _last = params['trend_segments'][-1]
    _last_end = _last.get('end', length - 1)
    if _last_end < length - 1:
        trend[_last_end + 1:] = trend[_last_end]
    seasonality_type = params.get('seasonality_type', None)
    seasonality = np.zeros(length)
    def _build_seasonal(t_arr, s_type, period, amplitude, phase, harmonics, envelope_type, length_seg):
        seg = np.zeros(len(t_arr))
        if s_type is None or s_type == 'none' or amplitude == 0:
            return seg
        t_local = np.arange(len(t_arr), dtype=float)
        envelope = np.linspace(0.7, 1.3, len(t_arr))
        omega = 2.0 * np.pi * t_arr / period + phase
        seg = amplitude * np.cos(omega)
        for h in harmonics:
            order = h.get('order', 2)
            amp_ratio = h.get('amplitude_ratio', h.get('relative_amplitude', 0.0))
            seg += amplitude * amp_ratio * np.cos(order * 2.0 * np.pi * t_arr / period + phase)
        return seg * envelope

    seasonal_regions = params.get('seasonal_regions', None)
    period = params.get('period', length)
    amplitude = params.get('amplitude', 0.0)
    phase = params.get('phase', 0.0)
    harmonics = params.get('harmonics', [])
    envelope_type = params.get('amplitude_envelope_type', 'constant')
    seasonality = _build_seasonal(t, seasonality_type, period, 
    amplitude, phase, harmonics, envelope_type, length)
    changes_array = np.zeros(length)
    for change in params.get('changes', []):
        ch_type = change.get('change_type', 'spike')
        direction = change.get('direction', 'up')
        tp = change.get('turning_point', length // 2)
        mag = abs(change.get('magnitude', 0.0))
        if direction == 'down':
            mag = -mag
        sp = change.get('start_position', None)
        ep = change.get('end_position', None)
        sigma = max(1.0, (ep - sp) / 4.0)
        changes_array += mag * np.exp(-0.5 * ((t - tp) / sigma) ** 2)
    noise_segments = params.get('noise_segments', None)
    noise_std = abs(params.get('noise_std', 0.0))
    noise = np.zeros(length)
    raw_ts = trend + seasonality + changes_array + noise
    target_mean = params.get('target_mean', None)
    target_std = params.get('target_std', None)
    raw_mean = raw_ts.mean()
    raw_std = raw_ts.std()
    ts = (raw_ts - raw_mean) / raw_std * target_std + target_mean
    return ts
\end{lstlisting}

\clearpage
\subsection{Case 3: ETTm1-96 HUFL}
\label{app:case_study_3}
\vspace{-0.8em}
{\fontsize{5.85}{5.1}\selectfont
\noindent\textbf{Text description.} This sequence is HUFL. The time series spans 96 points with mean 7.95, std 5.32, range [-8.97, 16.28]. Upward linear trend from baseline -3.08 to 12.91 from position 0 to 47. Downward linear trend from baseline 15.79 to 6.16 from position 48 to 95. No periodic/seasonal pattern. One spike+up from position 0 to 2 (peak at 1), magnitude 0.21. One spike+down from position 1 to 3 (peak at 2), magnitude 0.44. One spike+up from position 2 to 4 (peak at 3), magnitude 0.47. One spike+down from position 3 to 5 (peak at 4), magnitude 0.38. One spike+up from position 4 to 6 (peak at 5), magnitude 0.38. One spike+down from position 6 to 8 (peak at 7), magnitude 0.21. One spike+up from position 7 to 9 (peak at 8), magnitude 0.21. One spike+down from position 12 to 14 (peak at 13), magnitude 0.33. One smooth+up from position 17 to 25 (peak at 21), magnitude 0.17. One smooth+down from position 30 to 42 (peak at 36), magnitude 0.17. One spike+up from position 45 to 47 (peak at 46), magnitude 0.17. One spike+down from position 66 to 68 (peak at 67), magnitude 0.17. One smooth+up from position 66 to 72 (peak at 69), magnitude 0.33. One smooth+down from position 72 to 76 (peak at 74), magnitude 0.15. One spike+up from position 79 to 81 (peak at 80), magnitude 0.15. One spike+down from position 80 to 82 (peak at 81), magnitude 0.37. One smooth+up from position 89 to 95 (peak at 92), magnitude 0.23. Noise std\_deviation 1.958.\par}
{\fontsize{6.2}{5.6}\selectfont
\noindent\textbf{Execution.} \(L=96\); 17 events; MSE 0.0061; DTW 0.0471; PC 0.9315.\par}
\vspace{-0.3em}
\begin{center}
\includegraphics[width=0.82\linewidth,height=0.105\textheight,trim=4pt 8pt 4pt 8pt,clip,keepaspectratio]{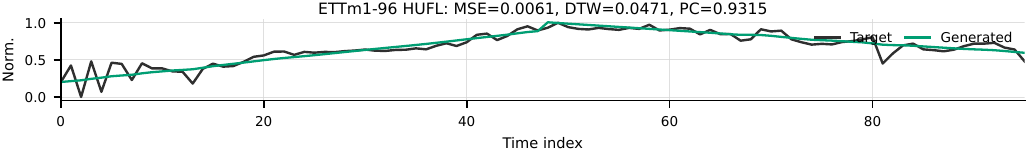}
\end{center}
\vspace{-1.5em}
\noindent\textbf{Folded params.}
\begin{lstlisting}[breaklines=true,breakatwhitespace=false,columns=fullflexible,basicstyle=\ttfamily\fontsize{5.9}{6.1}\selectfont]
length=96; target_mean=7.952; target_std=5.324; noise_std=1.9584
trend: linear(0-47): -3.084 -> 12.914; linear(48-95): 15.794 -> 6.164
seasonality: none
events: 17 total = 5 smooth + 12 spike; 10 with magnitude >= 0.2; max magnitude=0.474
top events: spike+up at 3 [2,4], mag=0.474; spike+down at 2 [1,3], mag=0.444; spike+down at 4 [3,5], mag=0.384; spike+up at 5 [4,6], mag=0.384
\end{lstlisting}
\vspace{-0.5em}
\noindent\textbf{Generated code.}
\begin{lstlisting}[language=Python,breaklines=true,breakatwhitespace=false,columns=fullflexible,basicstyle=\fontsize{5.9}{6.1}\ttfamily]
def generate_ts(params):
    import numpy as np
    length = params['length']
    t = np.arange(length, dtype=float)

    def _trend_seg(n, sv, ev, stype, extra):
        if n <= 0:
            return np.array([])
        if n == 1:
            return np.array([sv])
        tn = np.arange(n, dtype=float) / (n - 1)
        return sv + (ev - sv) * tn

    trend = np.zeros(length)
    for _s in params['trend_segments']:
        _a, _b = (_s.get('start', 0), _s.get('end', length - 1))
        trend[_a:_b + 1] = _trend_seg(_b - _a + 1, _s.get('start_value', 0.0), _s.get('end_value', _s.get('start_value', 0.0)), _s.get('trend_type', 'constant'), _s)
    _last = params['trend_segments'][-1]
    _last_end = _last.get('end', length - 1)
    if _last_end < length - 1:
        trend[_last_end + 1:] = trend[_last_end]
    seasonality = np.zeros(length)
    changes_array = np.zeros(length)
    for change in params.get('changes', []):
        ch_type = change.get('change_type', 'spike')
        direction = change.get('direction', 'up')
        tp = change.get('turning_point', length // 2)
        mag = abs(change.get('magnitude', 0.0))
        if direction == 'down':
            mag = -mag
        sp = change.get('start_position', None)
        ep = change.get('end_position', None)
        sigma = max(1.0, (ep - sp) / 4.0)
        changes_array += mag * np.exp(-0.5 * ((t - tp) / sigma) ** 2)
    noise_segments = params.get('noise_segments', None)
    noise = np.zeros(length)
    noise_std = abs(params.get('noise_std', 0.0))
    raw_ts = trend + seasonality + changes_array + noise
    raw_mean = raw_ts.mean()
    raw_std = raw_ts.std()
    ts = (raw_ts - raw_mean) / raw_std * params.get('target_std', raw_std) + params.get('target_mean', raw_mean)
    return ts
\end{lstlisting}

\clearpage

\newpage
\section*{NeurIPS Paper Checklist}

\begin{enumerate}

\item {\bf Claims}
    \item[] Question: Do the main claims made in the abstract and introduction accurately reflect the paper's contributions and scope?
    \item[] Answer: \answerYes{} % Replace by \answerYes{}, \answerNo{}, or \answerNA{}.
    \item[] Justification: The abstract and introduction state the main contributions of CodeTS, including the Text-to-Code-to-TS formulation, normalized executable code interface, synthetic Text-Code-TS triplet initialization, and execution-based RLVR optimization. The stated experimental claims are supported by comparisons on eight public benchmarks across three generation lengths, and the scope is qualified by the limitations discussion on univariate generation and ambiguous descriptions.

\item {\bf Limitations}
    \item[] Question: Does the paper discuss the limitations of the work performed by the authors?
    \item[] Answer: \answerYes{} % Replace by \answerYes{}, \answerNo{}, or \answerNA{}.
    \item[] Justification: The conclusion discusses the main scope limitations, including the focus on univariate time series and sensitivity to ambiguous descriptions. Additional risks related to synthetic data usage and executable code generation are discussed in the broader impact section.

\item {\bf Theory assumptions and proofs}
    \item[] Question: For each theoretical result, does the paper provide the full set of assumptions and a complete (and correct) proof?
    \item[] Answer: \answerNA{} % Replace by \answerYes{}, \answerNo{}, or \answerNA{}.
    \item[] Justification: The paper does not present theoretical results, theorems, or formal proofs. The mathematical formulations are used to define the Text-to-Code-to-TS framework, reward design, and optimization objective.

    \item {\bf Experimental result reproducibility}
    \item[] Question: Does the paper fully disclose all the information needed to reproduce the main experimental results of the paper to the extent that it affects the main claims and/or conclusions of the paper (regardless of whether the code and data are provided or not)?
    \item[] Answer: \answerYes{} % Replace by \answerYes{}, \answerNo{}, or \answerNA{}.
    \item[] Justification: The paper provides the key information needed to reproduce the main results, including benchmark datasets, metrics, model initialization, reward definitions, prompts, and SFT/RLVR hyperparameters. An anonymized repository is provided to support code reproduction.

\item {\bf Open access to data and code}
    \item[] Question: Does the paper provide open access to the data and code, with sufficient instructions to faithfully reproduce the main experimental results, as described in supplemental material?
    \item[] Answer: \answerYes{} % Replace by \answerYes{}, \answerNo{}, or \answerNA{}.
    \item[] Justification: An anonymized code repository is provided in the paper, and the appendix documents the public data sources used for training and evaluation. The repository provides the implementation and supporting materials for reproducing the main experimental results.

\item {\bf Experimental setting/details}
    \item[] Question: Does the paper specify all the training and test details (e.g., data splits, hyperparameters, how they were chosen, type of optimizer) necessary to understand the results?
    \item[] Answer: \answerYes{} % Replace by \answerYes{}, \answerNo{}, or \answerNA{}.
    \item[] Justification: The paper specifies the benchmark datasets, generation lengths, evaluation metrics, baseline categories, model backbone, SFT and RLVR training settings, reward weights, and optimization hyperparameters. Additional details on reward definitions, prompts, real data sources, and training configuration are provided in the appendix.

\item {\bf Experiment statistical significance}
    \item[] Question: Does the paper report error bars suitably and correctly defined or other appropriate information about the statistical significance of the experiments?
    \item[] Answer: \answerYes{} % Replace by \answerYes{}, \answerNo{}, or \answerNA{}.
    \item[] Justification: The paper reports aggregate performance metrics, including MSE, DTW, and Pearson correlation, to evaluate the generated time series against the ground-truth series.

\item {\bf Experiments compute resources}
    \item[] Question: For each experiment, does the paper provide sufficient information on the computer resources (type of compute workers, memory, time of execution) needed to reproduce the experiments?
    \item[] Answer: \answerNo{} % Replace by \answerYes{}, \answerNo{}, or \answerNA{}.
    \item[] Justification: The paper reports the main compute hardware and memory configuration, including the use of NVIDIA A100-SXM4-80GB GPUs for SFT and RLVR training. However, it does not provide wall-clock running time, per-experiment execution time, or an estimate of total compute consumption.
    
\item {\bf Code of ethics}
    \item[] Question: Does the research conducted in the paper conform, in every respect, with the NeurIPS Code of Ethics \url{https://neurips.cc/public/EthicsGuidelines}?
    \item[] Answer: \answerYes{} % Replace by \answerYes{}, \answerNo{}, or \answerNA{}.
    \item[] Justification: The research uses public datasets and pretrained models, reports the evaluation protocol, and discusses limitations, broader impacts, and safeguards for executable-code generation. To the best of the authors' knowledge, the work conforms to the NeurIPS Code of Ethics.

\item {\bf Broader impacts}
    \item[] Question: Does the paper discuss both potential positive societal impacts and negative societal impacts of the work performed?
    \item[] Answer: \answerYes{} % Replace by \answerYes{}, \answerNo{}, or \answerNA{}.
    \item[] Justification: The appendix includes a Broader Impact section discussing potential positive impacts in data-scarce time series domains such as healthcare, energy, transportation, finance, and climate analysis. It also discusses negative risks, including over-reliance on synthetic data, inherited bias, misleading downstream evaluations, and security concerns from executable code generation.
    
\item {\bf Safeguards}
    \item[] Question: Does the paper describe safeguards that have been put in place for responsible release of data or models that have a high risk for misuse (e.g., pre-trained language models, image generators, or scraped datasets)?
    \item[] Answer: \answerYes{} % Replace by \answerYes{}, \answerNo{}, or \answerNA{}.
    \item[] Justification: The paper describes safeguards for executable code generation, including normalized code formatting, format validation, sandboxed execution, and success rate reporting. The broader impact section further notes that real world deployment should include domain-specific validation, provenance tracking, privacy assessment, and disclosure of synthetic data usage.

\item {\bf Licenses for existing assets}
    \item[] Question: Are the creators or original owners of assets (e.g., code, data, models), used in the paper, properly credited and are the license and terms of use explicitly mentioned and properly respected?
    \item[] Answer: \answerNo{} % Replace by \answerYes{}, \answerNo{}, or \answerNA{}.
    \item[] Justification: The paper credits the main datasets, pretrained models, and software components through citations and public source URLs, and the appendix lists the public RLVR training data sources. However, it does not yet explicitly document the license names, versions, and terms of use for all existing assets.

\item {\bf New assets}
    \item[] Question: Are new assets introduced in the paper well documented and is the documentation provided alongside the assets?
    \item[] Answer: \answerYes{} % Replace by \answerYes{}, \answerNo{}, or \answerNA{}.
    \item[] Justification: The paper provides an anonymized repository containing the implementation and supporting materials for CodeTS. The paper and appendix document the training procedure, reward design, prompts, data sources, limitations, and broader impact considerations relevant to these released assets.

\item {\bf Crowdsourcing and research with human subjects}
    \item[] Question: For crowdsourcing experiments and research with human subjects, does the paper include the full text of instructions given to participants and screenshots, if applicable, as well as details about compensation (if any)? 
    \item[] Answer: \answerNA{} % Replace by \answerYes{}, \answerNo{}, or \answerNA{}.
    \item[] Justification: The paper does not involve crowdsourcing experiments or research with human subjects. All experiments are conducted on public time series datasets, synthetic triplets, and model generated outputs.

\item {\bf Institutional review board (IRB) approvals or equivalent for research with human subjects}
    \item[] Question: Does the paper describe potential risks incurred by study participants, whether such risks were disclosed to the subjects, and whether Institutional Review Board (IRB) approvals (or an equivalent approval/review based on the requirements of your country or institution) were obtained?
    \item[] Answer: \answerNA{} % Replace by \answerYes{}, \answerNo{}, or \answerNA{}.
    \item[] Justification: The paper does not involve crowdsourcing or human-subjects research, so IRB approval or equivalent review is not applicable.

\item {\bf Declaration of LLM usage}
    \item[] Question: Does the paper describe the usage of LLMs if it is an important, original, or non-standard component of the core methods in this research? Note that if the LLM is used only for writing, editing, or formatting purposes and does \emph{not} impact the core methodology, scientific rigor, or originality of the research, declaration is not required.
    %this research? 
    \item[] Answer: \answerYes{} % Replace by \answerYes{}, \answerNo{}, or \answerNA{}.
    \item[] Justification: LLMs are a core methodological component of the work: CodeTS is initialized from Qwen2.5-Coder-7B-Instruct and trained to generate executable code from textual time series descriptions. The paper describes this use in the method, experimental setup, training configuration, and prompt appendix.

\end{enumerate}

\end{document}